\documentclass[final,5p,times,twocolumn]{elsarticle}

\usepackage{amsmath}

\usepackage{color}
\usepackage{xspace}
\usepackage{pdfwidgets}
\usepackage{enumerate}
\usepackage{subcaption}
\def\ttdefault{cmtt}

\usepackage{xcolor}

\usepackage{amssymb}
\usepackage{tabulary}
\usepackage{algorithm}
\usepackage{algpseudocode}
\usepackage{natbib}
\usepackage{graphicx}

\journal{Neural Networks}

\newtheorem{definition}{Definition}

\usepackage{lineno}

\begin{document}

\begin{frontmatter}



\title{Concept Learning in the Wild: Towards Algorithmic Understanding of Neural Networks}



\author[bgu]{Elad Shoham} 
\ead{shellad@post.bgu.ac.il}
\author[bgu]{Hadar Cohen}
\ead{chada@post.bgu.ac.il	}
\author[bgu]{Khalil Wattad}
\ead{wattad.khalil@gmail.com}
\author[aca]{Havana Rika}
\ead{havanari@mta.ac.il	}
\author[bgu]{Dan Vilenchik}
\ead{vilenchi@bgu.ac.il}
\affiliation[bgu]{organization={Ben-Gurion University of the Negev},
            }
\affiliation[aca]{organization={The Academic College of Tel Aviv-Yaffo},
            }

\begin{abstract}
Explainable AI (XAI) methods typically focus on identifying essential input features or more abstract concepts for tasks like image or text classification. However, for algorithmic tasks like combinatorial optimization, these concepts may depend not only on the input but also on the current state of the network, like in the graph neural networks (GNN) case. This work studies concept learning for an existing GNN model trained to solve Boolean satisfiability (SAT). 
\textcolor{black}{Our analysis reveals that the model learns key concepts matching those guiding human-designed SAT heuristics, particularly the notion of 'support.' We demonstrate that these concepts are encoded in the top principal components (PCs) of the embedding's covariance matrix, allowing for unsupervised discovery. Using sparse PCA, we establish the minimality of these concepts and show their teachability through a simplified GNN. Two direct applications of our framework are (a) We improve the convergence time of the classical WalkSAT algorithm and (b) We use the discovered concepts to ``reverse-engineer"  the black-box GNN and rewrite it as a white-box textbook algorithm.
Our results highlight the potential of concept learning in understanding and enhancing algorithmic neural networks for combinatorial optimization tasks.}

\end{abstract}



\end{frontmatter}

\section{Introduction} \label{SEC: intro}

AI systems are typically thought of as computational entities that perform tasks already within the realm of human abilities but with enhanced efficiency and reduced effort. Nevertheless, the AI system may use new concepts that supplement human knowledge while performing a task. But even if concepts are already known, humans' trust in AI decisions increases when the mode of operation is transparent. Thus, having a framework that explains the underlying principles that guide the AI system's decisions is helpful.  This field of research is broadly called XAI (explainable AI).

One prominent line of research in XAI focuses on different aspects of the feature selection problem. Methods like SHAP
\cite{lundberg2017unified}, LIME \cite{lime},  GradCAM \cite{selvaraju2017grad}, or others \cite{Dorar, JungO21} provide insight into which features or patterns in the input lead to a particular output. \textcolor{black}{These methods provide a clear cause-and-effect explanation for input data where the set of features is pre-defined and named (e.g. in tabloid data or pixel location in an image).} However, systematically summarizing and manually interpreting such per-sample feature importance scores is challenging. Thus, instead of algorithms that assign importance to individual pixels in an input image,  we would like to have a concept-learning algorithm that, for example, identifies the police logo as an essential concept for detecting police vans.

Recent work studied ways to provide a concept-based explanation, which goes beyond per-sample features to identify higher-level human-understandable concepts that apply across the entire dataset \cite{yeh2020completeness,ghorbani2019towards, achtibat2022towards,alvarez2018towards,schut2023bridging,
kim2018interpretability,bau2017network}. Concepts are typically represented as vectors in the latent space of the neural network (NN). The concepts are found automatically, usually as the output of some optimization problem considering the labeled data and the specific NN. For example, one may consider a concept important if its removal from input images degrades classification performance \cite{ghorbani2019towards}.

One restriction of the works mentioned above is that they apply to concepts encoded entirely in the input data (say, in tasks like image or text classification). However, consider a graph neural network (GNN) that solves an algorithmic task, such as finding a satisfying assignment of a \textcolor{black}{$k$-CNF (Conjunctive Normal Form)} formula or finding a $k$-coloring of a graph. Such networks operate in a message-passing manner and may maintain a current solution that is being refined from iteration to iteration. In this setting, the concepts that the GNN learns need not depend solely on the input (e.g., the graph or the CNF formula), but on the interaction between the input and the ``working memory" of the algorithm where the current solution is stored. Furthermore, the concepts are dynamic and may evolve gradually with the message-passing procedure. This is in contrast to a single pass-forward of a CNN or a transformer that classifies images.

Work that finds concepts for algorithmic tasks includes \cite{schut2023bridging} that analyze AlphaZero \cite{silver2018general} playing chess. A new class of chess moves was discovered in \cite{schut2023bridging}
that were not observed in human-played games. {The method uses convex optimization to find key vector embedding, which later, with expert interpretation, is converted to the concept. However, playing chess doesn't require additional working memory as the board encodes all the information there is (plus a single bit of whose turn it is). Other methods, such as \cite{nanda2023progress}, analyze an NN that computes $(a+b) \mod p$. Although its mechanistic methodology gives a clear numerical cause-and-effect understanding, it is still difficult to achieve a human-explainable explanation.}

\begin{figure*}[t!]
    \centering
    \includegraphics[width=5in]{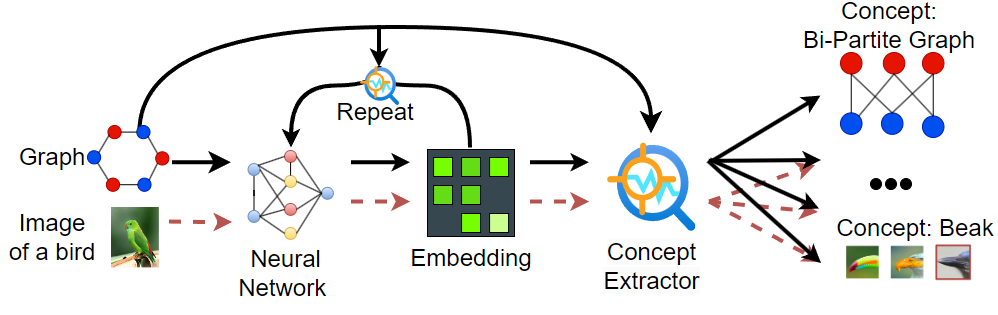}
    \caption{Red dashed arrows show the standard concept learning pipeline, input is embedded, and concepts are extracted\textcolor{black}{, where the concept of "Beak" is learned for image processing}. The black arrows show our setting, where input participates in concept learning, together with the embedding, through a dynamic process of repeatedly applying the NN to solve the algorithmic task, \textcolor{black}{for instance, the concept of bipartiteness}. The "Concept: Beak" image was taken from \cite{cao21}.
     }
    \label{FGR: pipeline}
\end{figure*}

In this work, we extend the concept-learning paradigm in a post-hoc \cite{posthoc} manner to combinatorial optimization problems, which comprise a central set of tasks in computer science. We focus on the problem of \textcolor{black}{Boolean} satisfiability (SAT) and study a popular GNN called NeuroSAT that was trained to solve SAT \cite{NeuroSAT19}. Figure \ref{FGR: pipeline} illustrates this work's pipeline \textcolor{black}{with an example of easily comprehensible concepts. The theoretical concepts are more intricate and will be elaborated upon later, focusing exclusively on Boolean satisfiability - SAT.}

\textcolor{black}{Our approach aligns more closely with the goal of concept-based cognition, where understanding a system's behavior involves identifying essential concepts that help humans (or algorithms, in our case) form rules and make decisions \cite{GUO2024102426}. Indeed, research in this direction shows that the concepts presented in this work share similar attributes with cognitive concept learning (CCL) as mentioned in \cite{GUO2024102426}: annotation (i.e., name of concept) and connotation (i.e., the definition of concept) in the attempt to fulfill the abstract-machine-brain characteristics of CCL. This is done by statistical analysis to emphasize the significant existence of concepts. Yet, out}
work crucially departs from previous work in that a concept needs to depend not only on the input to the NN but also on additional working memory. It is dynamic, learned, and evolves in repeated applications of the NN. In SAT, for example, an algorithm looks at the input $F$ (the CNF formula) and the current assignment $\phi$ that the algorithm will update in the next iteration if $\phi$ is not a satisfying assignment of $F$. The algorithm may compute various statistics such as: $(a)$ The number of variables in $F$ that appear only once (this concept depends only on the input $F$); $(b)$ The number of unsatisfied clauses in $F$ under $\phi$ (this depends on $F$ and $\phi$); $(c)$ For each variable $x$, the number of clauses that will become unsatisfied if the assignment of $x$ is flipped in $\phi$ (again, depends on $F,\phi$). Such concepts and many others may underpin the operation of the NN. 

There are several possible definitions of a concept; following \cite{schut2023bridging}, we define concepts as $(a)$ a unit that contains knowledge, i.e., information that is useful to solve a task; $(b)$ we can teach an algorithm or person the concept (transfer of the knowledge). If the person or algorithm grasps the concept, they can use it to solve the task. $(c)$ minimality -- redundant information is removed.  

\textcolor{black}{In this work, we focus on concept learning in a neural network model that achieves concept understanding without any architectural or loss function modifications to assist the learning process. This is a distinct approach from previous works, such as \cite{semi-sup-GCP, scattering, usup-GCP}, where the input to the GNN model is augmented with hand-crafted features or the loss function is modified to encourage specific behaviors. In contrast to these approaches, NeuroSAT operates without manually engineered features or specialized loss interventions, relying solely on the network’s internal learning dynamics. 
Our contribution lies in demonstrating how a model trained on a combinatorial optimization task can be interpreted through an algorithmic lens, providing a pathway to explainability. This insight not only sheds light on the network's problem-solving mechanisms but also paves the way for practical improvements to the GNN model and may inspire the creation or augmentation of hand-crafted algorithms. In essence, we are exploring 'concept learning in the wild,' observing the neural network’s emergent understanding solely based on exposure to the problem itself, without any externally imposed guidance or hints.
This is showcased with the following:}

\begin{itemize}
\item We show that NeuroSAT learns numerous algorithmic concepts that guided humans in the design of SAT heuristics before the advent of ML. Specifically, the central concept of {\em support} (\textcolor{black}{$(b)$} above) which was used in works like \cite{flaxman2003spectral,flaxman2008algorithms,achlioptas2008algorithmic}; Detail in Section \ref{sec:concepts}.
\item The concepts are encoded in the leading two principal components (PCs) of the covariance matrix of the latent space embedding of the NN. \textcolor{black}{The explained variance of the two leading PCs is above $95\%$}. Therefore, concepts are discovered in an unsupervised manner, unlike most work, where the class variable is used for concept mining. \textcolor{black}{Further details in Section \ref{sec:concepts}} 
\item We show minimality using the notion of sparsity (similarly to \cite{schut2023bridging}) by using a sparse PCA algorithm \textcolor{black}{to reduce the dimension of the embedding. Detailed in Section \ref{sec:concepts}}.
\item We show that the concepts are teachable by constructing a simple message-passing GNN and showing that it can be trained in a teacher-student manner to solve SAT, achieving the same performance as NeuroSAT, \textcolor{black}{and using 91\% less weights and biases}. The training minimizes the loss of concept learning rather than predicting SAT.  See Section \ref{sec:techaer_student}. 
\item A direct application of our approach is to improve a hand-crafted classical SAT algorithm, the WalkSAT algorithm \cite{walksat}. We updated its flipping rule according to one of the concepts we identified for NeuroSAT \textcolor{black}{to reach convergence $\approx 1.5$ times faster}. See Section \ref{sec: textbook}.
\end{itemize}

Finally, we use the aforementioned concepts in two downstream XAI tasks. The first is to rewrite the black-box NN as a white-box algorithm. We take all the concepts learned and use them to rewrite NeuroSAT as a textbook algorithm (see Section \ref{sec: textbook}). 
Second, we use the concepts to simplify the architecture of NeuroSAT, for example, replacing the LSTM with a much simpler RNN, and find that the performance remains the same.


The paper proceeds with a self-contained description of the SAT problem and relevant terminology in Section \ref{SEC: preliminaries}. Sections \ref{SEC: related work} and \ref{SEC: NeuroSAT model} survey related work, and in detail the NeuroSAT algorithm. Section \ref{SEC: data} describes the data that we've used in this paper, and Sections \ref{sec:concepts}--\ref{sec: textbook} detail our results. Concluding remarks are given in Section \ref{SEC: conclusion}.

\section{SAT Concepts} \label{SEC: preliminaries}
We start with the definition of SAT and then review the key algorithmic concepts referred to throughout the paper. 

\begin{definition} (CNF Formula)
A Boolean formula $F$ is in $k$-CNF form if it is the conjunction (logical $\land$) of $m$ clauses over $n$ variables. Each clause is of the form $(\ell_{1} \vee \ell_{2} \vee \cdots \vee\ell_k)$; each literal $\ell_i$ is a variable $x_i$ or its negation $\bar{x}_i$. 
\end{definition}

We denote by $c=m/n$ the \emph{density} of the formula, which is the ratio of constraints to variables. For clarity, we stick to the canonical form with $k=3$. 

An assignment $\phi:\{0,1\}^n \to \{0,1\}$ satisfies $F$  if it satisfies every clause in $F$, that is, in every clause, at least one literal evaluates to $True$ under $\phi$. The $k$-SAT decision problem is to decide whether a given $k$-CNF formula has a satisfying assignment $\phi$. The search problem is to find one such $\phi$  or declare that $F$ is unsatisfiable.

\begin{definition} (Random and Planted SAT) \label{defn:rand}
An instance from the \emph{Random 3SAT} distribution ${\mathcal{R}}(m,n)$ is sampled by randomly 
picking $m$ clauses out of \textcolor{black}{all $2^3 
\binom{n}{3}$ possible clauses configurations over $n$ variables (first, $3$ variables are chosen from the $n$ variables, $\binom{n}{3}$ ways for that, and then $2^3$ ways to set their literal form - negated or not).}

A \emph{Planted 3SAT} instance in ${\mathcal{P}}(m,n)$ is generated by first picking an assignment $\phi$ and then picking $m$ random clauses that are satisfied by $\phi$. 
\end{definition} 
Random SAT instances undergo a SAT-UNSAT phase transition as the density increases. The threshold is sharp as $n \to \infty$ and is located around $c=4.26$  \cite{PT:2003}. On the other hand, Planted SAT instances are satisfiable by definition for every $c$.

\begin{definition} (the concept of Support)\label{defn:support}
We say that a variable $x$ \emph{supports} a clause $C = (\ell_x \vee \ell_y \vee \ell_z)$ with respect to an assignment $\phi$ if its literal $\ell_x$ evaluates to $True$ and $\ell_y,\ell_z$ to False. That is, $x$ is the only one to satisfy that clause.
\end{definition}

To illustrate support suppose that  $C=(x \vee \bar{y} \vee \bar{z})$ and $\phi(x)=\phi(y)=\phi(z)=True$, then $\phi(C)=(T \vee F \vee F)$, thus  $x$ supports $C$ under $\phi$. Note that the supporting variable (if there is one) is unique for every clause.

The size of the support of $x$ naturally encodes a certain level of confidence in $\phi(x)$. Suppose that $x$ supports $s$ clauses. Consider $\phi'(x)$ in which the assignment of $x$ is flipped. All $s$ clauses in which $x$ appears are unsatisfied under $\phi'$ and may be corrected by flipping other variables. Therefore, large support $s$ implies that either $x$ is correctly assigned or, if incorrectly, then $\phi$ is incorrect on possibly $s$ additional variables. The latter is less likely when $\phi$ is close to a satisfying assignment.

\begin{definition} (the concept of WalkSAT)\label{defn:Walksat}
WalkSAT \cite{walksat,walksat_proof}  is a heuristic that starts from a random assignment and updates it by flipping the assignment of variables in unsatisfied clauses. Below is the basic skeleton of the algorithm that runs in time $O((n+m) \cdot T)$, where $T$ is the maximum number of iterations. 
\end{definition}

\begin{algorithm}
\caption{WalkSAT \cite{walksat,walksat_proof}}
\begin{algorithmic}[1]
\State \textbf{Input:} CNF Boolean formula $F$ with $n$ variables and $m$ clauses
\State \textbf{Output:} Satisfying assignment (if found) or UNSAT

\State Initialize $\varphi_0 \in \{ 0,1\}^n$ to a random assignment
\For{$i = 0$ \textbf{to} $T$}
    \State Evaluate $F$ under $\varphi_i$
    \If{$\varphi_i$ satisfies $F$}
        \State Return $\varphi_i$
    \EndIf
    \State Pick an UNSAT clause $c$ in $F$ uniformly at random 
    \State and a random variable in $c$, say $x_j$\Comment{$O(m)$}
     \State Set $\varphi_{i+1}$ to be $\varphi_i$ but with the assignment of $x_j$ flipped.  \Comment{$O(n)$}
\EndFor
\State \Return UNSAT
\end{algorithmic}
\end{algorithm}

There are different variants of WalkSAT; for example, the variable $x_j$ is flipped with probability $p$ ($p$ is a predefined parameter), and with probability $1-p$, a random flip occurs (or in a different version, the variable with least support is flipped).

\section{Related Work} \label{SEC: related work}

\begin{table*}
    \centering
    \begin{tabular}{| c |c |c |c |c |c |} 
     \hline
     Dataset & \# Instances & $n$ & $c$  & Source Name& Source \\ [0.5ex] 
     \hline
     SPARSE & $2800$ & $500,1000,1500,2000$ & $0.5,0.8,1,1.25,1.3,1.5,1.6$ & Random SAT &${\mathcal{R}}(m,n)$ \\  
     \hline
     DENSE & $2000$ & $500,1000,1500,2000$ & $3.75,4.0, 4.1, 4.2, 4.25$  & Random SAT&  ${\mathcal{R}}(m,n)$ \\ 
     \hline
    PLANTED & $3500$ & $500,1000,1500,2000$ & from $4.5$ to $15.5$ jumps of $0.5$ & Planted SAT &    ${\mathcal{P}}(m,n)$ \\ 
    \hline
    \end{tabular}
    \caption{Datasets used for concept learning of NeuroSAT. For each combination of $n$ and $c$, there are 100 random instances.}
    \label{table: datasets}
\end{table*}

\subsection{Concept Learning}
\textcolor{black}{
Concept-based methods have emerged as a powerful approach to model interpretability, offering a higher level of abstraction compared to traditional feature and data-centric interpretability methods \cite{sundararajan2017axiomatic,lundberg2017unified,lime}. These methods aim to provide model explanations that are more intuitive and informative for human practitioners. The core principle of concept excavation involves defining an optimization program over the set of possible concepts, typically represented as vectors in the neural network's latent space, and the available data.
}

\textcolor{black}{
Research in this field has branched into several distinct approaches. Supervised concept excavation \cite{achtibat2022towards,alvarez2018towards,schut2023bridging,kim2018interpretability,bau2017network} relies on labeled data to identify and validate concepts. In contrast, unsupervised concept mining \cite{yeh2020completeness,ghandeharioun2021dissect,ghorbani2019towards} seeks to discover inherent concepts without the need for labeled data. Some studies have explored concept learning within reinforcement learning frameworks \cite{SHI201947}, integrating the dynamic nature of RL environments with concept discovery.}

\textcolor{black}{A closely related paradigm, known as ``Mechanistic Interpretability'' or ``Mechanistic Design,'' takes inspiration from reverse engineering compiled binary code. This approach aims to deconstruct neural networks to identify specific functionalities, such as computing the Max function, by examining individual components of the model \cite{Transformers-circute,quantization,interpretability-gpt,zoom,nanda2023progress}.}

\textcolor{black}{Recent work has also extended concept learning to GNN \cite{RakaraddiLPC22,SunLZ22,GonzalezHC02}, 
focusing on identifying patterns in input data for structural learning or prerequisite information. Our work differentiates itself by not only examining patterns in the input data but also inspecting the intermediate memory of the model, providing a more comprehensive analysis of concept formation and utilization within neural networks.
}

\subsection{The SAT Problem}
In the previous section, we mentioned the classical hand-crafted SAT-solving heuristic WalkSAT. Alongside such polynomial time heuristics, exact solvers like zChaff \cite{mahajan2005zchaff2004} were introduced and used both in academia and industry. In recent years, neural network-based heuristics have also been developed for various combinatorial optimization problems such as  SAT. 
\cite{NeuroSAT19} propose an end-to-end NN-based approach called NeuroSAT. On the other hand, \cite{selsam2019guiding,yolcu2019learning} incorporate NN in classical search algorithms to better guide that search. The SAT-solving community has explored the application of machine learning to SAT in more practical contexts. Possibly the most successful example is SATzilla \cite{satzila}. 

Our work crucially departs from previous concept learning methods in our considered domain. While previous works assume that the concepts are imbued in the data, one should consider our dataset a set of Turing machines (in our case, a Turing machine for solving SAT). In this case, the concept depends on the machine's transition function as well as the working memory. This raises the level of complexity compared to finding concepts in textual data, images, or even in a chess game.

\section{The NeuroSAT Model} \label{SEC: NeuroSAT model}
\textcolor{black}{
CNNs, RNNs, and Dense models are less effective for combinatorial optimization problems. CNNs are primarily designed for grid-like data structures and rely on local connectivity \cite{cappart2023combinatorial}.  While RNNs can handle sequential data, they have limitations when dealing with graph-structured problems and face challenges with long-term dependencies and parallelization \cite{alomar2024rnns}. Fully connected or dense layers lack the ability to handle variable-sized inputs, which is a common characteristic of graph-structured data in combinatorial optimization problems. They also don't inherently capture the relational structure present in graphs. In contrast, GNNs are ideal for combinatorial optimization problems due to their ability to handle graph-structured data, capture complex relationships and dependencies, and propagate data from nodes to nodes regardless of the size of the graph \cite{cappart2023combinatorial,gasse2019exact}}.

NeuroSAT is a GNN running iteratively in a message-passing manner. The input to the GNN is a graph, as its name suggests. A CNF formula is readily described as a bipartite graph; one part consists of literals and the other of clauses. Edges connect literals to clauses in which they appear. This graph is called a factor graph (illustration in Figure \ref{FGR: sat_to_graph} in the appendix). 
\textcolor{black}{The NeuroSAT GNN consists of two key single-layer LSTMs $\mathbf{C_u}$ and $\mathbf{L_u}$ with hidden dimension 128  (the subscript $u$ stands for ``update") that compute the update rule of the aggregated clause-to-variable and variable-to-clause messages in each iteration. That computation also involves two MLPs $\mathbf{L_{msg},C_{msg}}$, and an operator $Flip(A)$ that takes a matrix indexed by the $2n$ literals and swaps each row $i$ with row $n+i$ (a literal and its negation).
The output of $\mathbf{C_u}$ in iteration $t+1$ is an $m \times 128$  matrix $C^{t+1}$ that packs the embedding of the $m$ clauses of the CNF input (with adjacency matrix $M$). It is computed from the clause embedding at the previous iteration, $C^t$, and the $2n\times 128$  variable embedding matrix $L^t$ using the following update rule:}
\begin{equation}\label{eq:ClauseUpdate}
    C^{t+1} \longleftarrow \mathbf{C_u}([C^t, M^\top \mathbf{L_{msg}}(L^t)]),
\end{equation}

\textcolor{black} {Similarly, the $2n\times 128$ literal embedding matrix $L^{t+1}$ is computed using the following update rule:}
\begin{equation}\label{eq:LiteralUpdate}
    L^{t+1} \longleftarrow \mathbf{L_u}([L^t, Flip(L^t), M \mathbf{C_{msg}}(C^t)]).
\end{equation}

NeuroSAT also has an initialization component; the full NN architecture can be found in the Appendix, Figure \ref{FGR: full_model}. 
    
The NN can be viewed as a message-passing algorithm with a parameterized number of iterations. The clause-to-variable messages are computed in one iteration using Eq.~\eqref{eq:LiteralUpdate}; all the variables-to-clause messages are computed in the next iteration using Eq.~\eqref{eq:ClauseUpdate}. 

NeuroSAT was trained in a supervised manner with a train set of millions of random CNF formulas with $n \in [10,40]$ variables, using the cross-entropy loss for predicting satisfiability.
The random formulas were chosen around the SAT-UNSAT threshold (at such small $n$'s, the threshold is not concentrated around 4.26 but rather lies in the interval $[4,6]$). An exact solver was used to label the formulas (hence the small $n$ value).
       
Alongside satisfiability prediction, a binary assignment to the variables can be computed from the literal embedding $L^t$. The original paper computed it using a simple clustering algorithm that partitions the literals' embedding into TRUE and FALSE.  
Figure \ref{FGR: clustering} illustrates this procedure.

\section{Dataset} \label{SEC: data}
This paper focuses on the task of SAT-solving rather than refutation (proving that a formula is UNSAT). Thus, we only consider satisfiable instances for the concept mining task. Table~\ref{table: datasets} describes the data we've used to demonstrate the concepts. 

{\color{black} {We selected our datasets to align with NeuroSAT's original training on Random 3SAT instances, aiming to analyze the specific concepts it learned within this context. By choosing different Random 3SAT datasets, each with distinct combinatorial properties, we can systematically explore which concepts NeuroSAT employs to address each dataset. Importantly, since NeuroSAT is known to struggle with general SAT instances (as the authors of the paper admit) —an outcome we also observed in preliminary experiments—we excluded them from our analysis to maintain a focused exploration of datasets where concepts are expected to be learned. However, we did challenge NeuroSAT by including a different distribution, the Planted SAT (see Definition \ref{defn:rand}), which was not part of NeuroSAT's training. PLANTED enables us to generate significantly larger instances in terms of variable count and density, allowing us to test how well NeuroSAT’s learned concepts generalize.}

We selected Random 3SAT datasets with varied densities identified in the literature to highlight significant combinatorial structure that also has algorithmic implications.} 
    The Random 3SAT datasets are partitioned into SPARSE and DENSE, representing different clause-to-variable ratios $c$. All $c$'s are below the SAT-UNSAT threshold.
It was widely observed in the rich random SAT literature that the value of $c$ governs the algorithmic difficulty of the problem. As $c$ increases and the formula becomes more constrained, structural changes occur in the topography of the SAT solution space (e.g., the 1RSB phase transition) \cite{rsa.1rsb,achlioptas2008algorithmic,krzakala2007gibbs}. Therefore, different concepts will be suitable for different density regimes. 

Another note is that we study formulas with thousands of variables compared to the tens used in training. This demonstrates the robustness of the concepts learned on minimal formulas and generalized to much larger ones.

The PLANTED dataset contains random satisfiable instances at ratios $c$ below and above the SAT-UNSAT threshold, as PLANTED instances are always satisfied.

\section{Evaluation}\label{sec:concepts}
In this section, we explain in detail what concepts have been learned by NeuroSAT.

We used a personal computer with an Nvidia RTX 3080 GPU, 64GB RAM, and an AMD Ryzen 9 5650x CPU to run the experiments. The code will be made public after it is accepted.

We ran NeuroSAT for 500 iterations on every input, whereas the original paper ran for 26 iterations, but we've seen that for more challenging instances, that was not enough.

Throughout this section, we refer to the first and second principal components, PC1 and PC2. The PCs are the two leading eigenvectors of the covariance matrix of either the clause or variable embedding (the $C^{t}$ and $L^{t}$ matrices in Eq.\eqref{eq:ClauseUpdate} and \eqref{eq:LiteralUpdate}). However, there are 500 hundred covariance matrices for $C^{t}$ and $L^{t}$, one for each iteration, repeated for every instance in the dataset. Since concepts are attributes of the distribution, not of a particular instance,  we decided to take the average over all instances and iterations and compute its PCs. More formally, for each of the three datasets $\mathcal{D}$ = SPARSE, DENSE, and PLANTED, we compute an average covariance matrix $S_\mathcal{D}$ as follows. For a CNF formula $F \in \mathcal{D}$, let $S_{F,i}$ be the covariance matrix of the literal (or clause) embedding when running for $F$ in iteration $i$ ($t=i$ in Eq.~\eqref{eq:LiteralUpdate}). Let 
\begin{equation}\label{eq:AvgCov}
    S_\mathcal{D} = \frac{1}{500  \lvert \mathcal{D} \rvert }\sum_{i=1}^{500} \sum_{F \in \mathcal{D}}  S_{F,i} 
\end{equation}

Finally, PC1 and PC2 are the top two eigenvectors of the matrix $S_\mathcal{D}$, explaining $\sim 98\%$ of the total variance (PC1 explains $\sim 90\%$ and PC2 explains $\sim 8\%$ ). 
Thus, the PCs are concepts of the distribution $\mathcal{D}$ similar to a beak concept of distribution over bird images.

To demonstrate the notion of concept minimality, we computed sparse versions of PC1 and PC2 by zeroing all but the top 16 entries (out of 128) with the largest absolute value. The number 16 was chosen by analyzing the distribution of the different embedding dimensions (details omitted). We use capitalized 'SPARSE' for the low-density random SAT formulas and 'Sparse' for sparse PCA.

We now proceed with the analysis of all the concepts that we identified for NeuroSAT.
We repeated the experiments in this section with the sparsified PC versions. The results were almost identical to the non-sparsified PCs. For brevity, we report the sparse results only for some concepts; complete results are in Tables \ref{tab:consistency sparse}--\ref{tab:appearance_count sparse} in the appendix. 

\subsection{The concept of assignment} \label{SEC: con_ass}
The most basic concept of a SAT-solving algorithm is a consistent Boolean assignment $\phi$. That is, every pair of literals $x_i$ and $\bar{x}_i$ receive the assignment $\phi(x_i,\bar{x}_i) = (True,False)$ or $\phi(x_i,\bar{x}_i)=(False,True)$ (but not both $True$ or both $False$, which we call a contradiction).

\begin{figure}
    \centering
    \includegraphics[width=1\columnwidth]{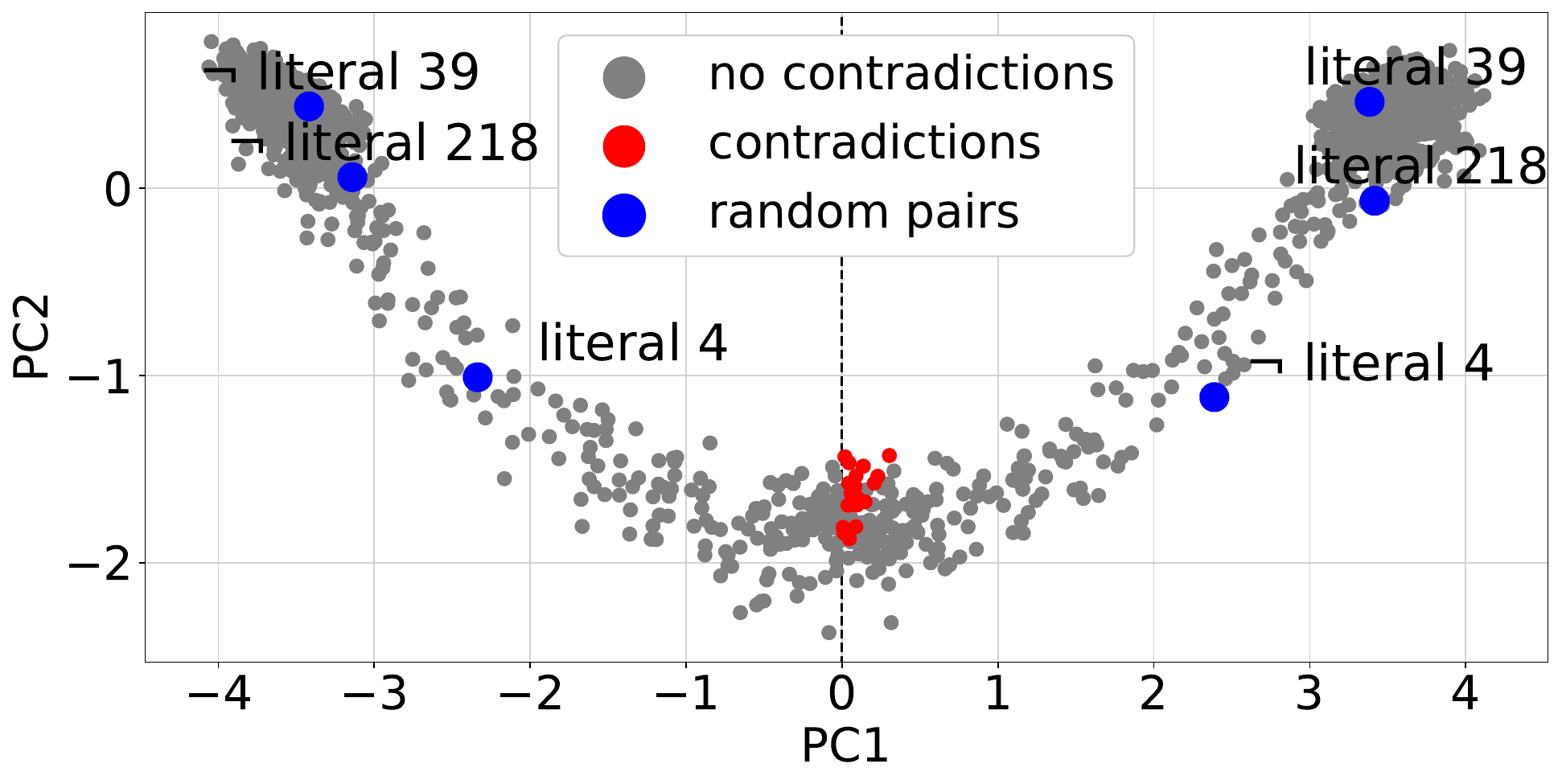}
    \caption{PCA projection of the literal embedding; embedding of a literal and its negation are symmetrical. Three random pairs of literals $(x_i,\bar{x_i})$ are colored with similar absolute PC2-values. Variables with contradiction are colored red and are all near PC1 equals 0.}
    \label{FGR: clustering}
    
\end{figure}

Figure \ref{FGR: clustering} plots the 2D PCA projection of NeuroSAT's literal embedding after the 500th iteration. We see a symmetric shape around $x=0$; most literals lie on opposite sides of $x=0$, roughly at the same $y$-value, inducing a consistent assignment for these literals.

Table \ref{tab:consistency} shows the percentage of contradictions for every dataset when assigning literals according to their PC1 value (positive vs negative). 

We observe slightly higher rates of contradiction for SPARSE instances, which come from variables that appear only in one polarity (thus, one form of literal negation is missing). The missing-form embedding is freely ``floating" on the board and may end up next to the appearing one.

\begin{table}
  \centering
  \begin{tabular}{|c| c| c|}
  \hline
     Dataset  & \multicolumn{2}{|c|}{\% of contradictions} \\
    \cline{2-3} 
    &  PC1 &  Sparse PC1   \\
    \hline
    SPARSE& 7.3 $\pm 2.7$ \% &6.4 $\pm 1.2$ \% \\ 
    \hline
     DENSE & 3.2 $\pm 1.5$ \% &  4.0 $\pm 1.0$ \%\\ 
    \hline
     PLANTED & 2.0 $\pm 1.0$\% & 2.0 $\pm 1.0$ \%\\ 
    
    \hline
  \end{tabular}
  \caption{High assignment consistency rates. The average \% of contradictions per iteration over all iterations and instances in the dataset. Left column uses PC1 values for clustering and right column the Sparse PC1.}
  \label{tab:consistency}
\end{table}

In what follows, we use $\phi_t$ for the assignment at iteration $t$. If $\phi_t$ contains inconsistencies, we solve them by assigning the literal with the larger PC1 value True and the other False.

After establishing the concept of assignment, it is natural to ask what is the \% of times that the assignment encoded by NeuroSAT at the 500th iteration was a satisfying one. Table \ref{table: performance} shows the rates.

In line with Random SAT literature, the instances get harder to solve as $c$ approaches the SAT-UNSAT threshold (asymptotically at $c=4.26$).

\begin{table}[t]
\centering
\begin{tabular}{|c |c |c |c |c |c |} 
 \hline
 Distribution / $c$ &  $\leq 3.8$  &  $4.0$ &  $4.1$ &  $4.2$ &  $4.25$   \\
 \hline
 RANDOM   & $100\%$  & $67\%$ & $48\%$ & $58\%$ &  $43\%$ \\
  PLANTED & $100\%$ & $100\%$ & $100\%$  & $100\%$  & $100\%$  \\
  \hline
\end{tabular}
\caption{The average rate of NeuroSAT finding a satisfying assignment. Rates drop as $c$ increases, in line with Random SAT theory.}
\label{table: performance}
\end{table}

\subsection{The concept of support} \label{SEC: con_support}
\begin{figure}
    \centering
    \includegraphics[ width=\columnwidth]{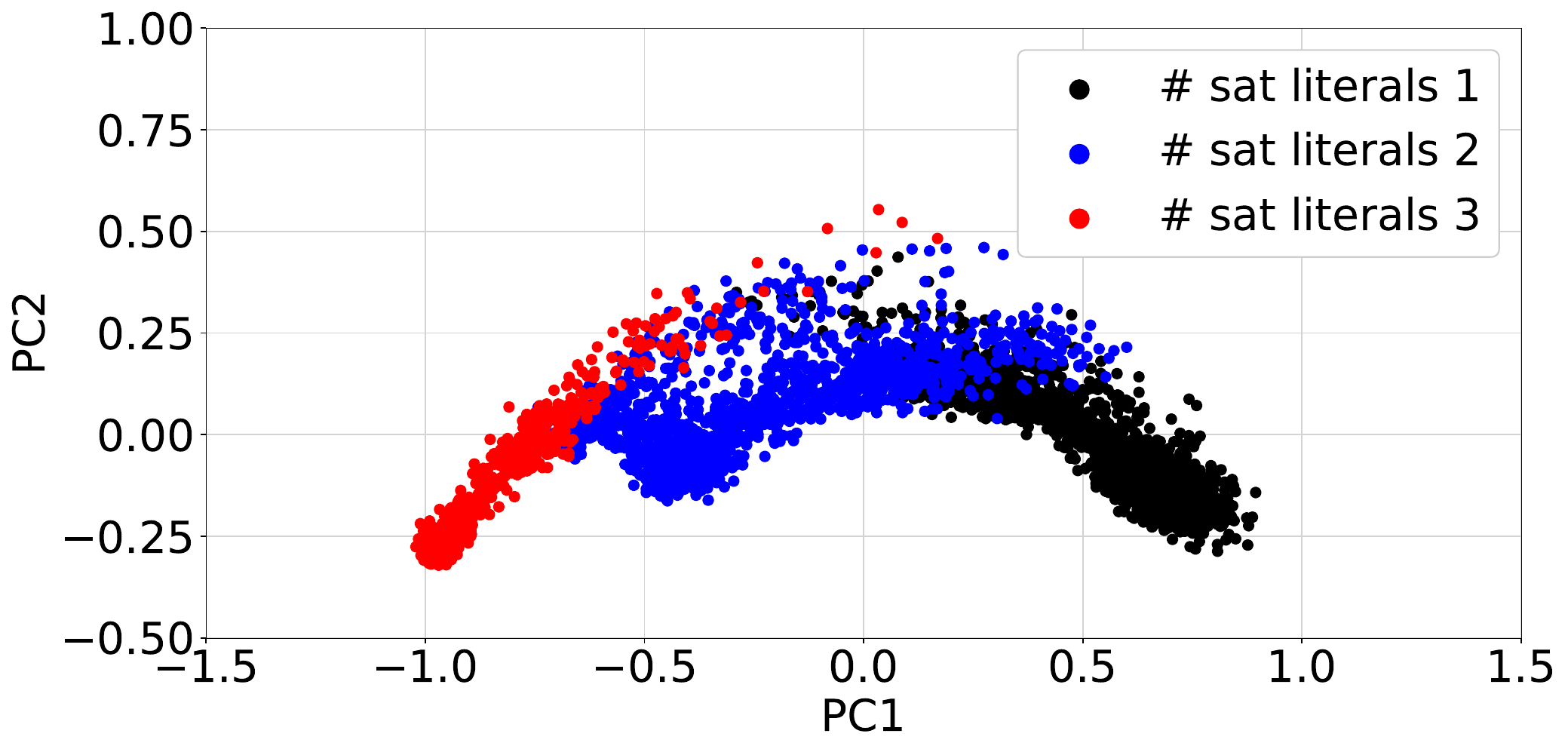}
    \caption{The clause embedding represents support. PCA projection of clause embedding for $c=3.75$ and $n=1500$ color-coded with the number of literals that satisfy the clause. Support clauses tend to have positive PC1 values.}
    \label{FGR: clause_embeddedings}
\end{figure}

A central concept in SAT-solving is the concept of support, recall Definition \ref{defn:support}. We now explore the full support-learning cycle of NeuroSAT. 
\\
\noindent \textbf{Clause embedding}: 
Every clause $C$ contains one, two, or three literals that evaluate to True under $\phi_t$.  Figure \ref{FGR: clause_embeddedings} plots the PCA projection of the clause embedding after iteration 500. As evident, the number of satisfied literals per clause is encoded by NeuroSAT, with a positive PC1 value being cut-off for supported clauses vs. nonsupported. We call a support clause with positive PC1 value a ``concept abiding" clause.

Table \ref{tab:Clause_support} gives the percentage of concept-abiding clauses with respect to $\phi_t$ (averaged over all iterations).   Note that for SPARSE instances, the rates are lower as, indeed, the concept is less pronounced in such low densities (a variable is expected to appear in $3c$ clauses, $3c/8$ of them count as support. This expectation exceeds 1 only at $c=8/3$, which is outside the SPARSE regime).

\begin{table}[h!]
  \centering
  \begin{tabular}{|c| c| c|}
    \hline
     Dataset & \multicolumn{2}{|c|}{\% of concept-abiding clauses} \\
    \cline{2-3} 
     &  PC1 &  Sparse PC1   \\
    \hline
    SPARSE & $61 \pm 8$ \% & $61 \pm 6$ \% \\ 
    \hline
     DENSE & $89 \pm 6$ \% & $87 \pm 7$ \% \\ 
    \hline
     PLANTED & $97 \pm 5$ \% & $97 \pm 5$ \% \\ 
   
    \hline
  \end{tabular}
  \caption{Clause support concept. The average \% of concept-abiding support clauses according to PC1 values and Sparse PC1. }
  \label{tab:Clause_support}
\end{table}

\begin{figure}[h!]
\centering
    \begin{subfigure}[t]{.95\columnwidth }
        \centering
         \includegraphics[ width=\textwidth]{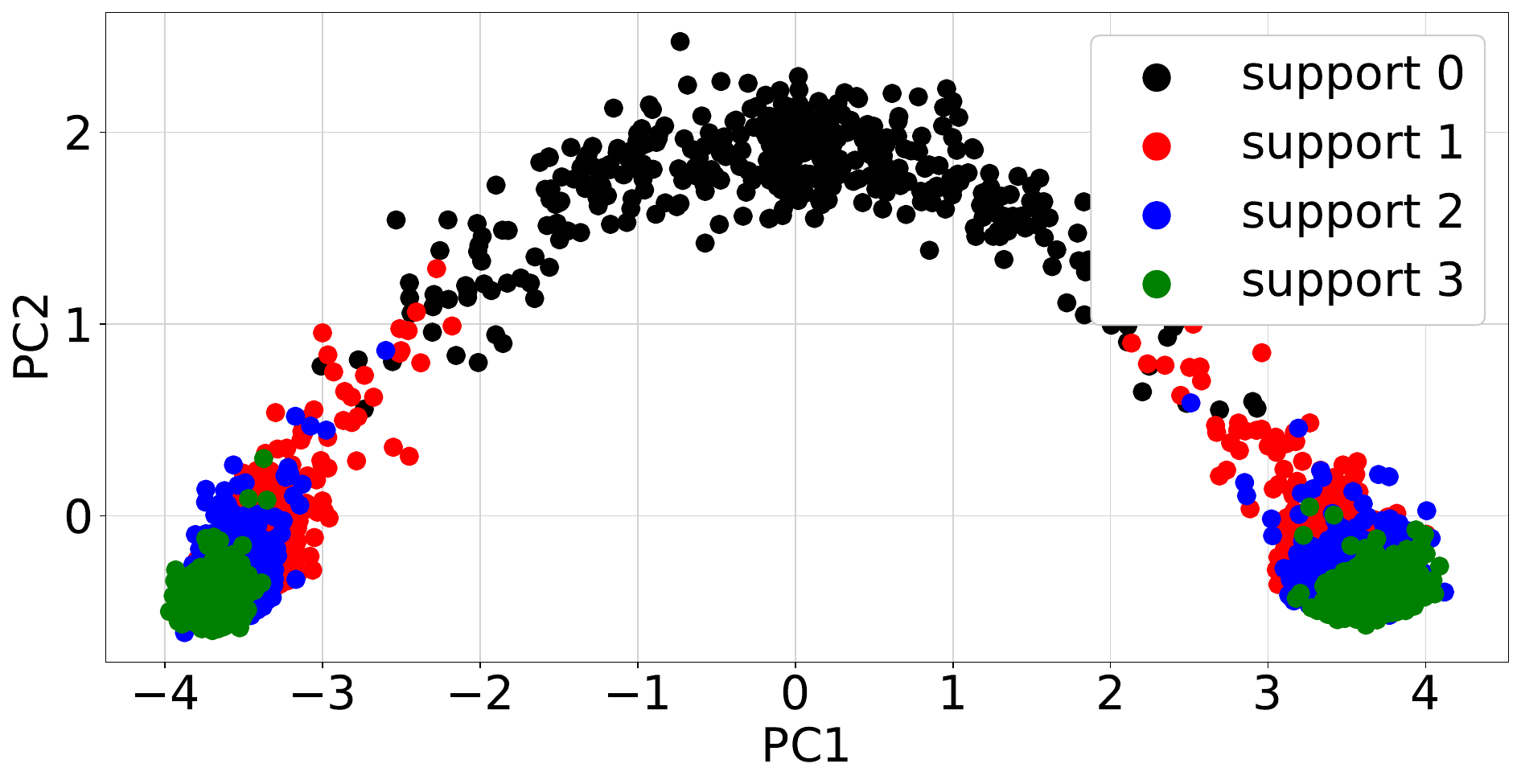}
         \subcaption{Support count - reaching a satisfying assignment}
         \label{FGR: support}
    \end{subfigure}
    \hfill
    \begin{subfigure}[t]{.95\columnwidth }
        \centering
         \includegraphics[ width=1\textwidth]{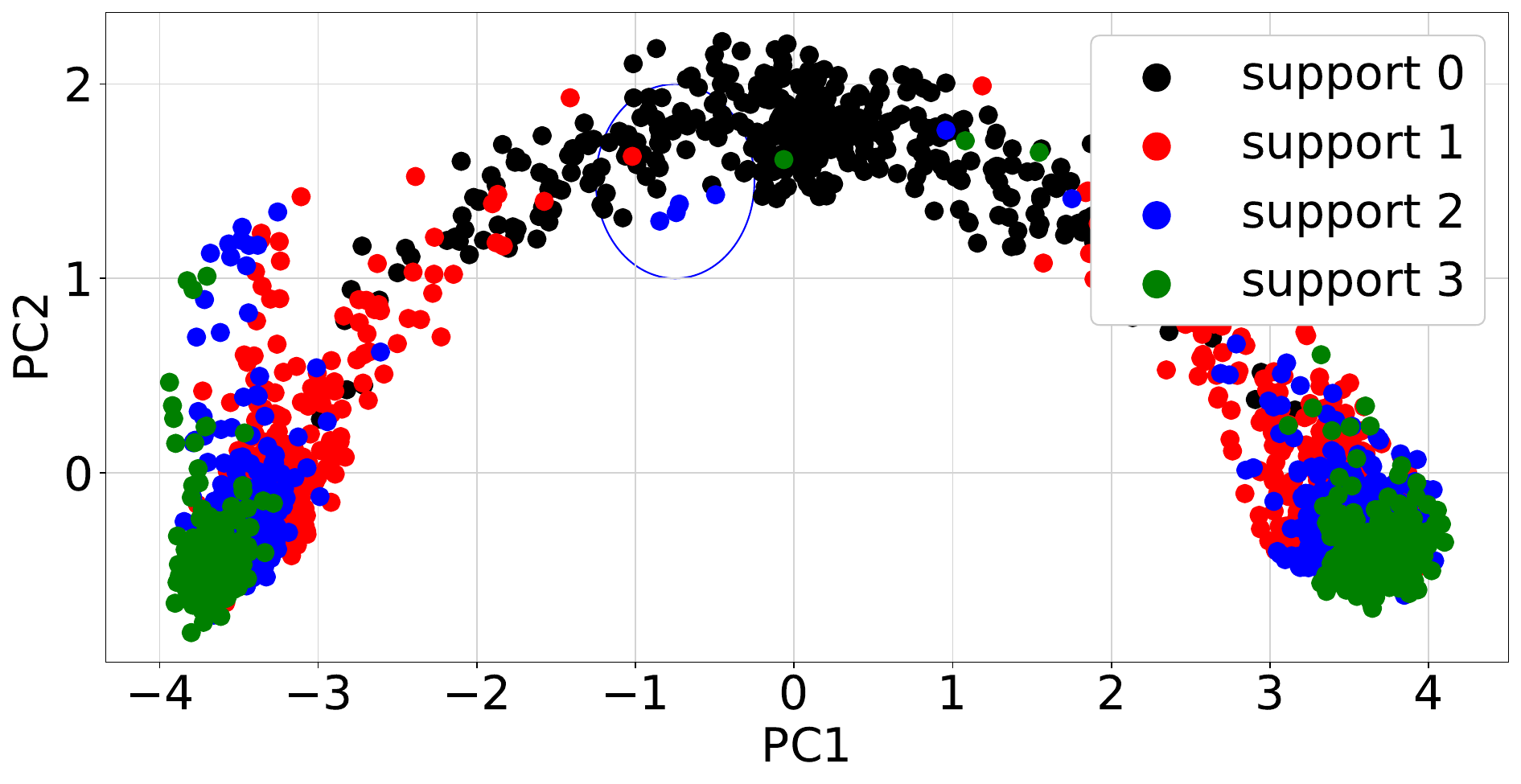}
        \subcaption{Support count - reaching a close, but unsatisfying assignment}
        \label{FGR: unsupport}
    \end{subfigure} 
    \hfill
    \begin{subfigure}[t]{.95\columnwidth }
        \centering
         \includegraphics[ width=1\textwidth]{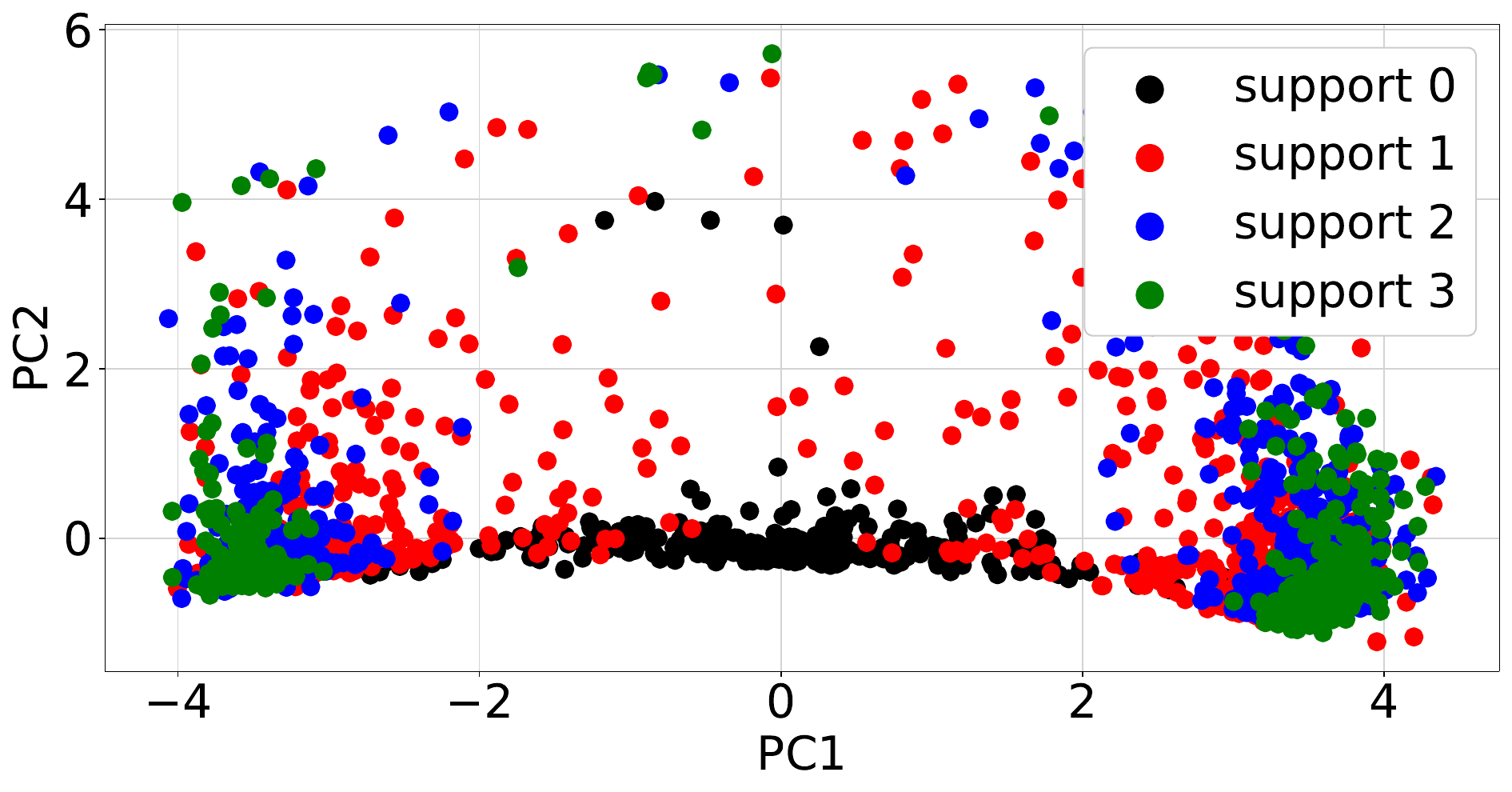}
        \subcaption{Support count - reaching a far unsatisfying assignment}
        \label{FGR: very-unsupport}
    \end{subfigure} 
    \caption{PCA projection of the literals' embedding for $c=4.1$ and $n=1500$. The embedding is color-coded with the support count; 
    A larger absolute $PC1$-value means larger support. Figures $(b),(c)$ are executions where NeuroSAT fails to find a satisfying assignment; support is encoded with noise.}
    \label{FGR: support_classification}
\end{figure}

\noindent\textbf{Literal embedding}:
The next piece in the support concept encoding is found in the literal embedding. The embedding of the clauses in which a literal appears is summed and fed into $\mathbf{L_u}$, the LSTM of the literal message, Eq.~\eqref{eq:LiteralUpdate}. Figure \ref{FGR: support} illustrates how the literal support is then encoded -- the larger the PC1 value (in absolute value), the larger the support. A  0-support zone is defined by PC1 values -2 and 2. In Section \ref{sec: textbook}, we emphasize the meaning of the $[-2,2]$ range, but for now, we simply note that this is a range of ``uncertainty", meaning that these are variables whose assignment can (or should) be flipped.

Table \ref{tab:Literal_support} shows the ranges of PC1 values for each support group and the percentage of clauses with that support that have a PC1 value in that range. Support 0 is well separated from the rest, while ranges increasingly overlap as the support value increases. In other words, NeuroSAT economizes on information and encodes ``low" vs. "high" support, not differentiating between the actual high values. As we'll explain shortly, this also makes algorithmic sense.

\begin{table}
  \centering
  \begin{tabular}{ |c |c |c |c |c |}
    \hline
    Dataset&Support0&Support1&Support2&Support3\\
    \hline
    SPARSE&$\pm[0, 2]$&$\pm[2.1, 3.3]$&$\pm[2.7, 3.5]$&$\pm[3, 3.7]$\\
    &94\%&80\%&86\%&80\%\\
    
    \hline
    DENSE&$\pm[0, 1.5]$&$\pm[1.6, 2.8]$&$\pm[2.4, 3.0]$&$\pm[2.6, 3.2]$\\
    &81\%&76\%&84\%&92\%\\
        \hline

   DENSE &  $\pm[0, 1.5]$&$\pm[1.6, 2.8]$&$\pm[2.4, 3.0]$&$\pm[2.6, 3.2]$\\
    FAILED&60\%&60\%&70\%&76\% \\
    \hline

   DENSE &$\pm[0, 1.5]$&$\pm[1.6, 2.8]$&$\pm[2.4, 3.0]$&$\pm[2.6, 3.2]$\\
    FAR &46\%&43\%&54\%&61\% \\
    \hline

  \end{tabular}
  \caption{Literal support concept. PC1 range and the percentage of clauses with support $i$ that fall in that range. We shorthand $\pm[a,b]$ for $[-a,-b] \cup [a,b].$ The first two rows correspond to successful executions of Neuorsat. The next row to a failed executions. The last row to a failed executions, and far from every satisfying assignment.}
  \label{tab:Literal_support}
\end{table}
 
\noindent\textbf{Failed executions}:
To highlight the necessity of the support concept, we show that when NeuroSAT fails to find a satisfying assignment, it also computes a poorer encoding of the support.
We differentiate between two types of failure: reaching close to a satisfying assignment (Hamming distance about 1\%), illustrated in Figure \ref{FGR: unsupport}, and far (distance $>$ 10\%), illustrated in Figure \ref{FGR: very-unsupport}.

This poorer encoding is also evident in the last two lines in Table \ref{tab:Literal_support}, which show a lesser consistency between the range and the actual support for the close-failed and even worse for far-failed. A crucial point is the misencoding of variables with large support, encoded by mistake as low support. As we shall see in Section \ref{sec: textbook}, NeuroSAT has difficulty setting them correctly.

The gradation in concept learning quality between successful vs. close-failed vs. far-failed executions shows how deeply interwoven this concept is in the NN's operation.

We now proceed to a more abstract concept that builds upon the concept of support: the concept of backbone variables.

\begin{figure*}[t!]
    \centering
    \begin{subfigure}[t]{1.0\columnwidth }
            \centering
            \includegraphics[ width=\columnwidth]{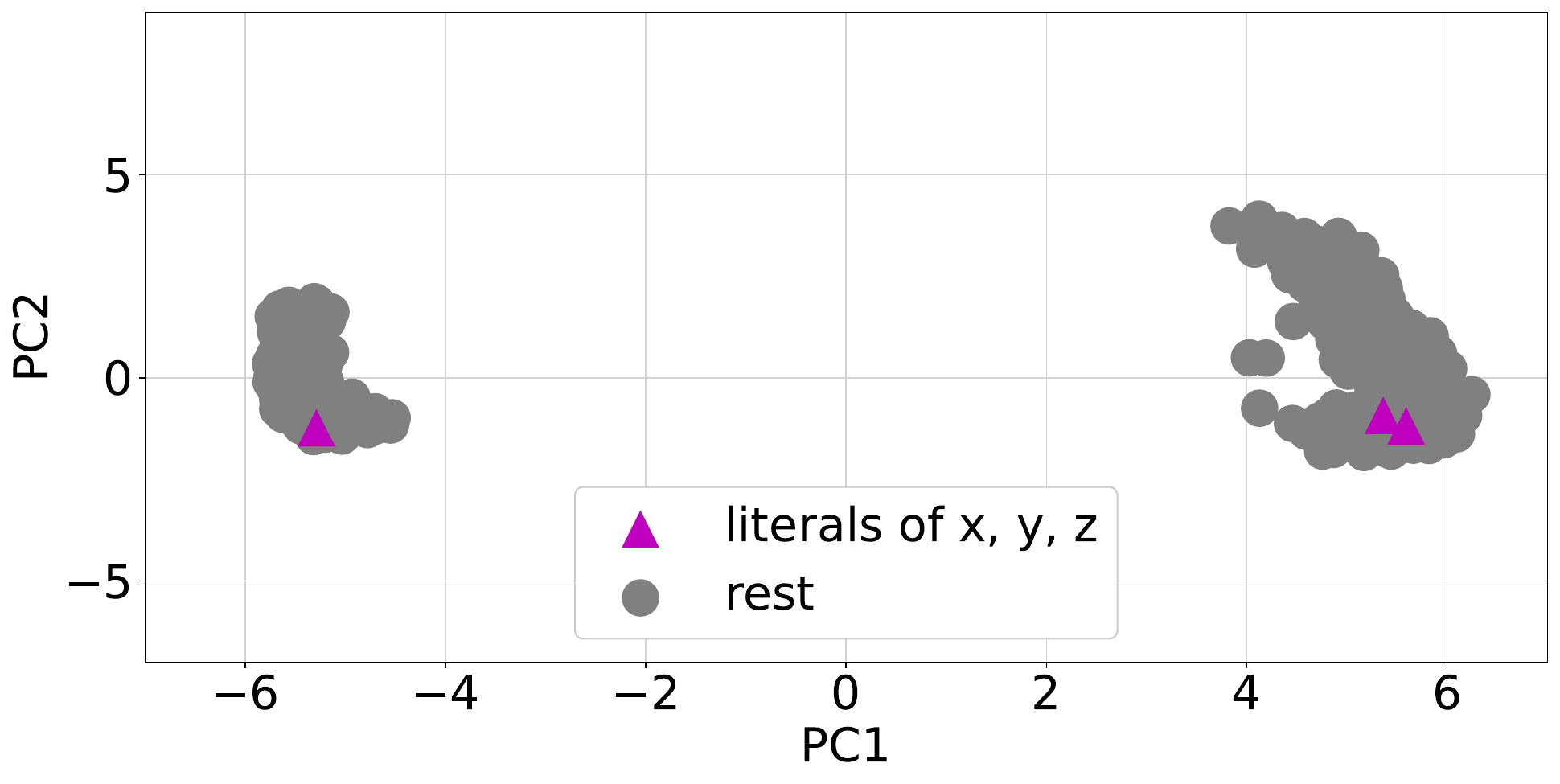}

            \subcaption{Valid assignment - backbone on PLANTED}
            \label{FGR: backbone_sat}
    \end{subfigure}
    \hfill
        \begin{subfigure}[t]{1.0\columnwidth }
            \centering
            \includegraphics[ width=\columnwidth]{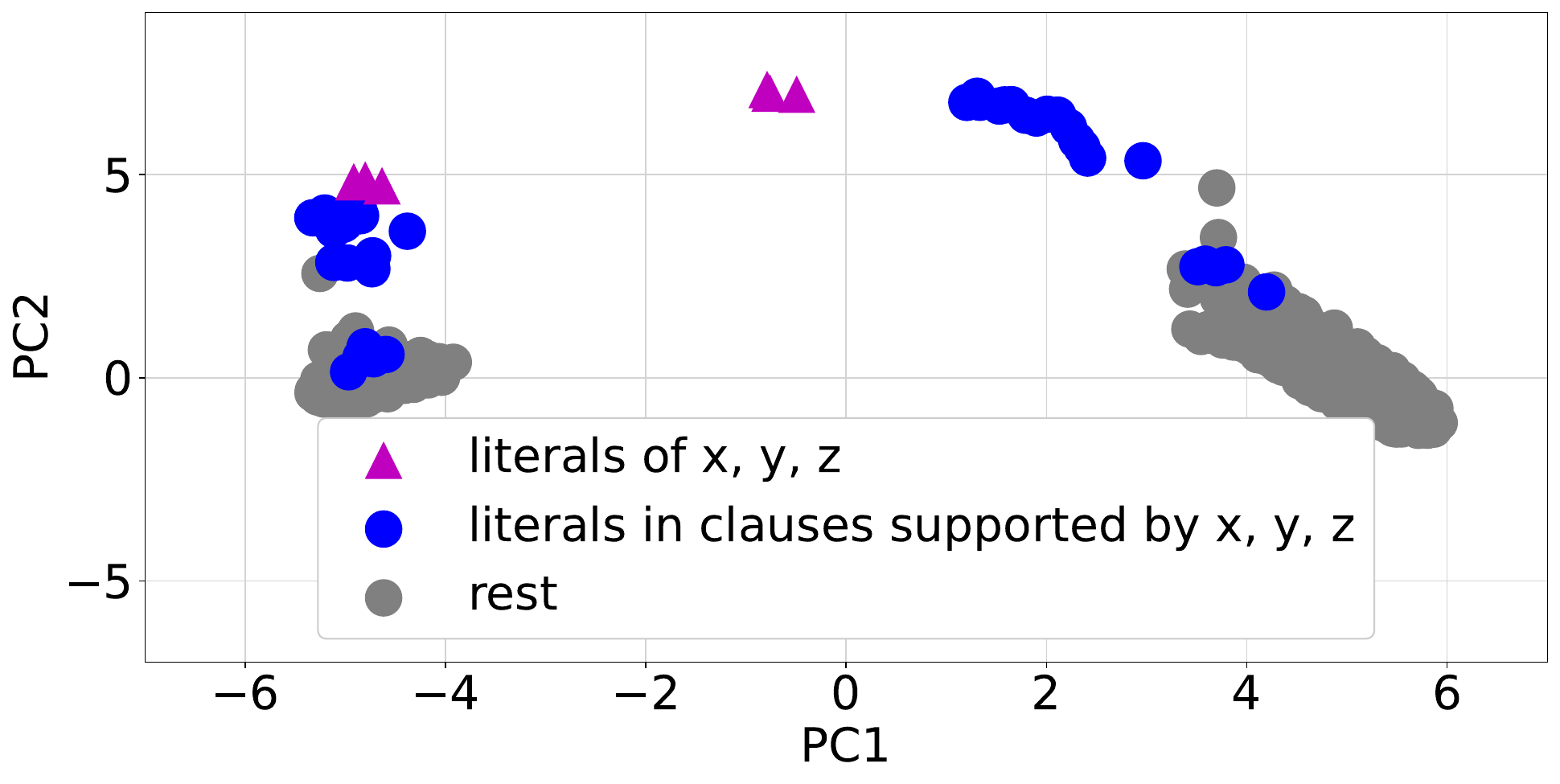}
            \subcaption{Invalid assignment - backbone on PLANTED}
            \label{FGR: backbone_unsat}
        \end{subfigure}
    \hfill
    \begin{subfigure}[t]{1.0\columnwidth }
            \centering
            \includegraphics[ width=\columnwidth]{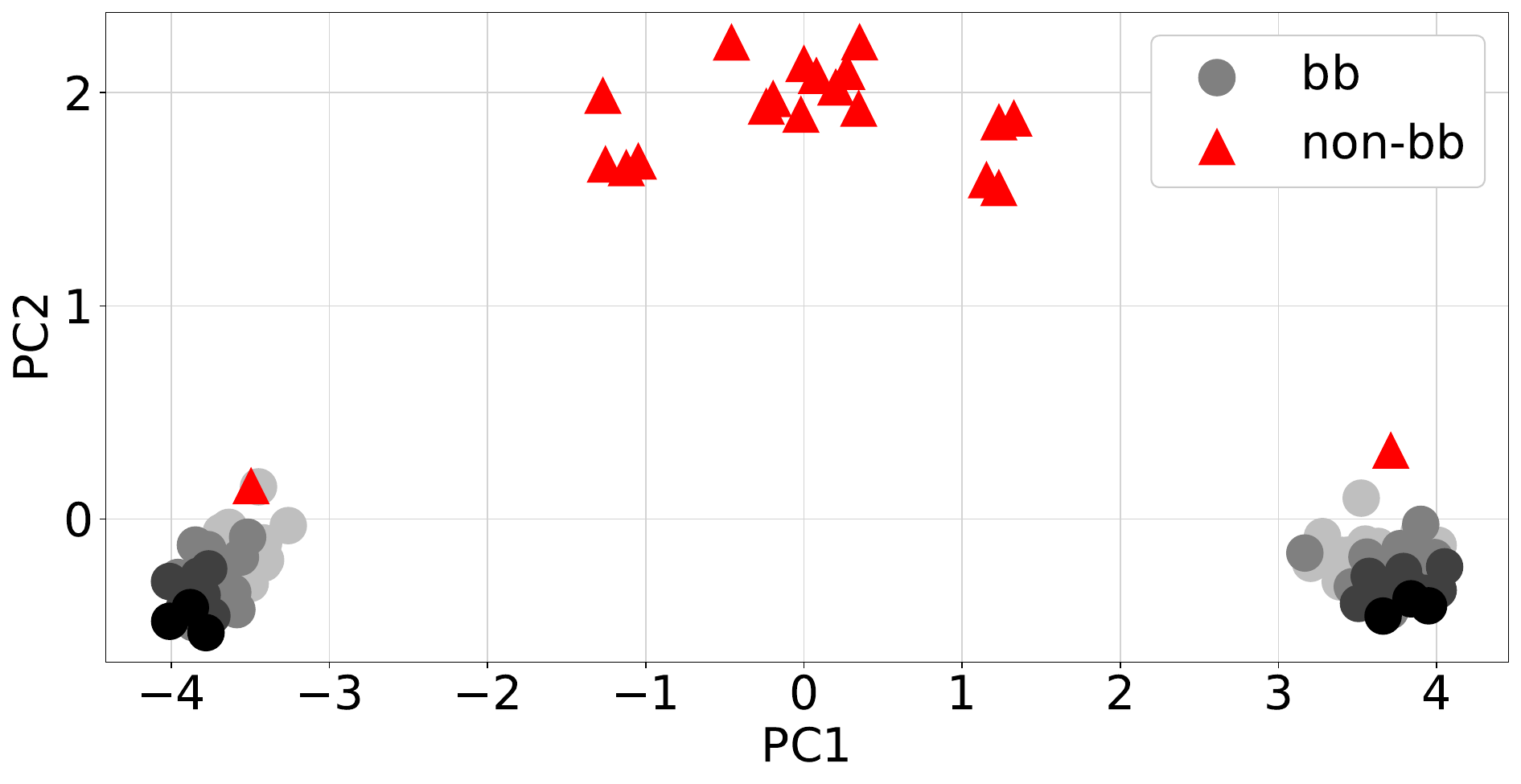}
            \subcaption{Valid assignment - backbone on SATLIB}

            \label{FGR: bb_satlib_valid}
        \end{subfigure}
    \hfill
        \begin{subfigure}[t]{1.0\columnwidth }
            \centering
            \includegraphics[ width=\columnwidth]{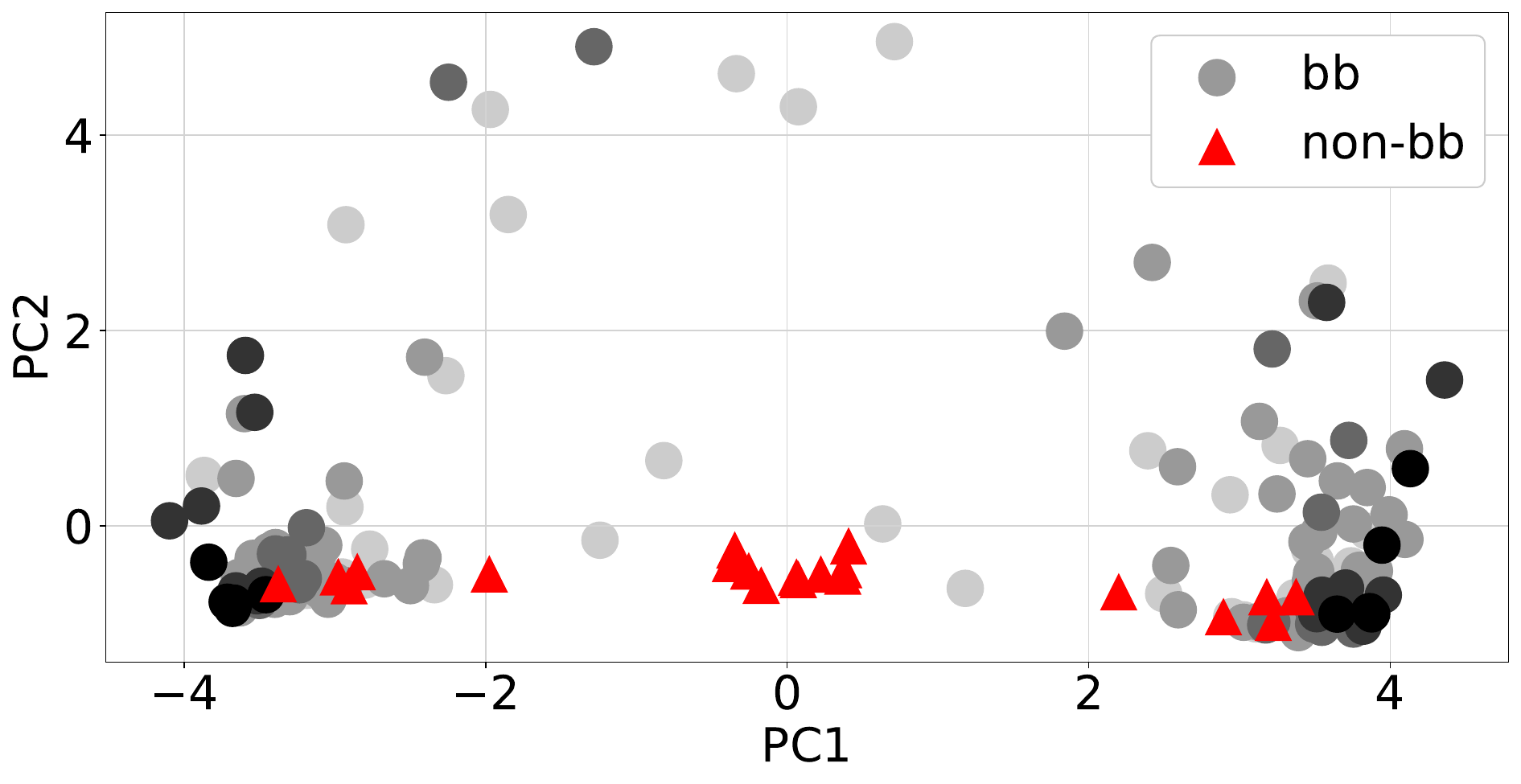}
            \subcaption{Invalid assignment - backbone on SATLIB}
            \label{FGR: bb_satlib__invalid}
    \end{subfigure}
\caption{ Figure \ref{FGR: backbone_sat} illustrates a clause $C$ containing three random non-backbone variables $x,y,z$ is added to a planted formula with $n=1000$ and $c=15$. Because $x,y,z$ are non-backbone variables, the formula remains satisfying, and no detection occurs. Figure \ref{FGR: backbone_unsat} is the same experiment, but this time $x,y,z$ are part of the backbone. Making the instance UNSAT as $x,y,z$ are all unsatisfied in $C$. NeuroSAT identifies the three.
Figure \ref{FGR: bb_satlib_valid} is NeuroSAT's embedding of the backbone dataset ($b=90\%$ from the SATLIB dataset). The black backbone variables are placed ``correctly'' outside the $[-2,2]$ interval of zero support, where the majority of non-backbone variables lie.
Figure \ref{FGR: bb_satlib__invalid} shows the embedding when NeuroSAT fails to reach a satisfying assignment; in this case, some backbone variables are misplaced. Backbone variables are also color-coded by support count, the darker the greyscale, the higher the support. 
} 
\label{FGR: backbone}
\end{figure*}

\subsection{The concept of backbone variables} \label{SEC: backbone}
The backbone of a CNF formula $F$ refers to the variables that take the same assignment in all satisfying assignments. E.g., if $x_1$ and $x_2$ are in the backbone of $F$, then, say, in all satisfying assignments of $F$, $x_1=True$ and $x_2=False$. These variables are sometimes referred to as frozen, but we will stick with the more common backbone. 

Dense PLANTED instances were shown to have such variables \cite{coja2007almost}, and in fact, when the density is $\Omega(\log n)$, there is typically only a single satisfying assignment. In \cite{achlioptas2008algorithmic}, it was shown that DENSE formulas, while not having a global backbone, have a local one with respect to a cluster of satisfying assignments. Different clusters may have different backbones assigned differently, even for shared variables. The proof in both cases relied heavily on the support concept.

Thus, we may consider backbone variables a more complex concept defined on top of the support. 

To show how NeuroSAT identifies backbone variables, we generate dense PLANTED instances with $c=15$. We then
use the efficient heuristic described in \cite{coja2007almost} to identify the backbone. We select three backbone variables at random, say $x,y,z$, and add a clause $C^*=(\bar{x} \vee \bar{y} \vee \bar{z})$ to $F$ (assume that the planted assignment $\phi$ is True for all three, otherwise change polarities accordingly). The resulting instance $F \cup C^*$ is UNSAT, but note that $C^*$ is merely one of $15n+1 = 15,001$ clauses of $F$ (for $n=1000$). We repeated the same experiment, now choosing three non-backbone variables; this time, $F \cup C^*$ remains SAT.

Figure \ref{FGR: backbone_sat} shows the typical dense PLANTED embedding, no longer with the arched form, as all literals have high support. The addition of $C^*$ has no effect as the formula remains satisfiable. In Figure \ref{FGR: backbone_unsat}, we see that NeuroSAT identifies the three backbone variables, ``saying" that they should have support 0 to avoid contradiction. It also identifies literals that appear in clauses that $x,y,z$ support.

We eyeballed 100 plots of instances with $n=1000$, $9<c<16$, and this pattern repeats consistently.

To further demonstrate the backbone concept, we used another source of CNF formulas, the well-known SATLIB benchmark library \cite{hoos2000satlib}. The library contains several datasets of ``backbone formulas," which are RANDOM formulas conditioned on having a backbone of size $b\%$, for values $b=10,30,50,70,90$, $n=100$, and densities around the SAT/UNSAT threshold ($c=4.25$). Using an exact solver, we found the backbone variables for instances with $b=30,50,70,90$. We ran NeuroSAT on 800 instances for each group. We found that when NeuroSAT finds a satisfying assignment, the PC1 values of backbone variables are never in the range $[-2,2]$ (Figure \ref{FGR: bb_satlib_valid}), while for instances that could not be solved by NeuroSAT, there is at least one backbone variable with a PC1 value in the range $[-2,2]$ (Figure \ref{FGR: bb_satlib__invalid}). Recall from Section \ref{SEC: con_support} that the range $[-2,2]$ is the 0-support zone (indeed, backbone variables, by definition, must have non-zero support with respect to every satisfying assignment). This strengthens the importance NeuroSAT gives to the backbone concept. Figure \ref{FGR: bb_satlib_valid} shows a valid assignment where all the backbone variables are not in the $[-2,2]$ range. Figure \ref{FGR: bb_satlib__invalid} shows an invalid assignment where we can see backbone variables in the $[-2,2]$ range.

Figures \ref{FGR: bb_satlib_valid} and \ref{FGR: bb_satlib__invalid} also show how the concept of support is applied to out-of-distribution formulas. Backbone formulas were not part of the Neurosat's training; however, as the grey-scale color-coding in those figures shows, PC1 encodes the support correctly for successful solution instances, and for failed solutions, it does not.

\subsection{The concept of majority vote}
For a variable $x$ in a CNF formula $F$, let $A_x^{(+)}$ be the set of clauses in which it appears as the literal $x$, and similarly, $A_x^{(-)}$ the set of clauses in which it appears as $\bar{x}$.

The Majority Vote assignment, MAJ(F), assigns every variable $x$ the value True if $\lvert A_x^{(+)} \rvert \ge \lvert A_x^{(-)} \rvert$ and False otherwise. MAJ is a popular starting point for SAT-solving heuristics, as it maximizes the total number of True literals in $F$ (counting every literal in every clause, with repetition).

Table \ref{tab:MAJ} shows the average distance between $\phi_1$, NeuroSAT's assignment after the first iteration, and MAJ. As evident from the table, NeuroSAT's starting point is MAJ.

\begin{table}[htb]
  \centering
  \begin{tabular}{ |c |c |c |c |}
    \hline
     Dataset  & \multicolumn{2}{|c|}{Hamming(MAJ,$\phi_1$)} \\
    \cline{2-3} 
    &   PC1 &  Sparse PC1   \\
    \hline 
     SPARSE &  $0.0 \pm 0.0$ & $0.14 \pm 0.03$ \\
    \hline
    DENSE & $0.0 \pm 0.0$ &   $0.20 \pm 0.013$ \\
    \hline
    PLANTED  & $0.00 \pm 0.00$  &  $0.17 \pm 0.014$  \\
    \hline
  \end{tabular}
  \caption{The Majority Vote concept. The average Hamming distance between MAJ and $\phi_1$, NeuroSAT's output after the 1st iteration derived from PC1 values (or sparse PC1), as explained in Figure \ref{FGR: clustering}.}
  \label{tab:MAJ}
\end{table}

\subsection{The concept of appearance count}
\begin{figure}[h!]
    \centering
    \includegraphics[ width=\columnwidth]{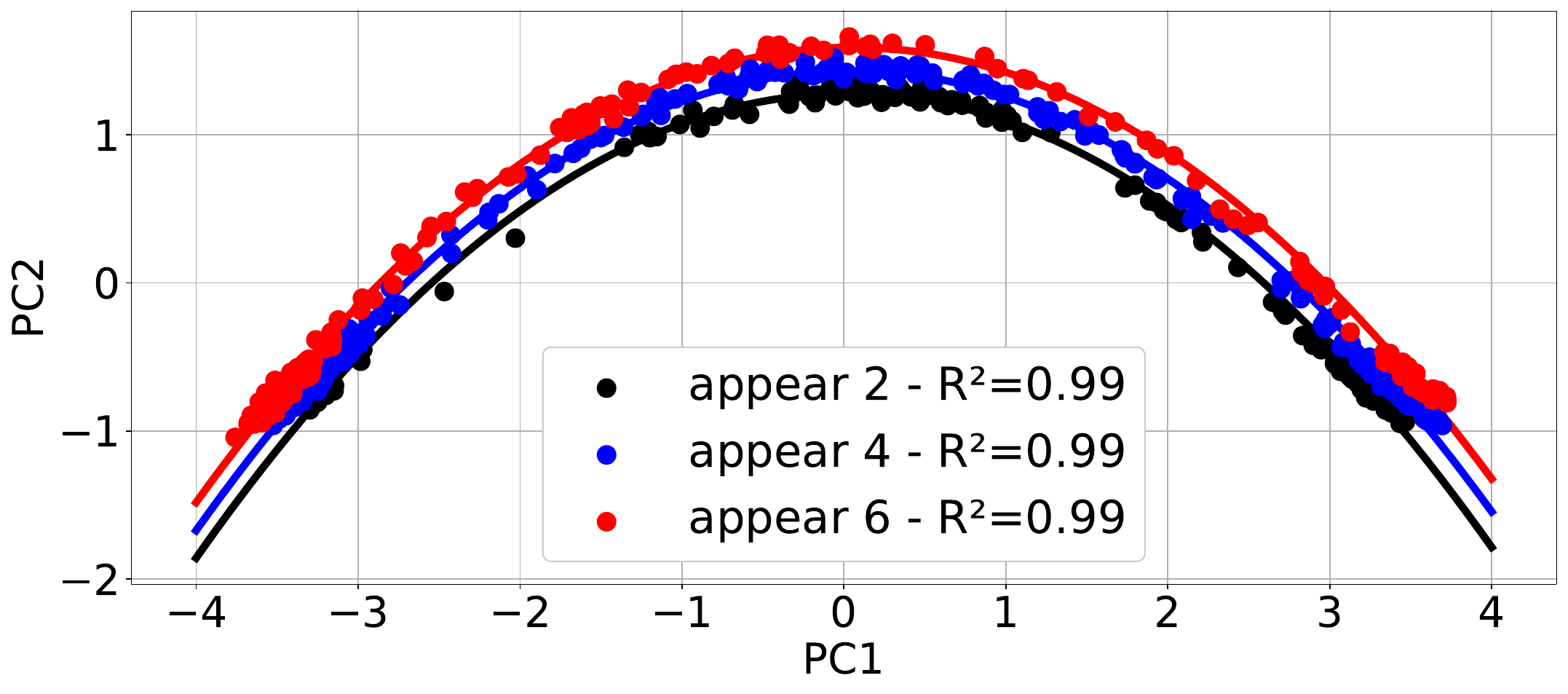}
    \caption{The appearance is distinctively separable. PCA projection of literal embedding color-coded by appearance count. Only showing the 2,4,6 appearance on the last (500th) iteration for $c=4.1,n=1500$. An arched structured by appearance cohorts.
    }
    \label{FGR:color_appearnce_plot}
\end{figure}

For a variable $x$ in a CNF formula $F$, $\lvert A_x^{(+)} \rvert + \lvert A_x^{(-)} \rvert$ is the total number of times that $x$ appears in $F$. Figure \ref{FGR:color_appearnce_plot} illustrates how NeuroSAT encodes the appearance count. Unlike previous concepts, which PC1 encoded, this concept is encoded by the parabolic shape spanned by PC1 and PC2.

Table \ref{tab:appearance_count} shows the percentage of variables that appear $i$ times that are closest to the polynomial regression line that fits all variables that appear $i$ times. Similarly to the support encoding, NeuroSAT does not differentiate between the exact number of appearances as they get large. Therefore we bucket $i=3,4,5$ and $i\ge 6$. In the bucketed columns, we report the \% of variables that appear $i$ times and are closest to one of the relevant regression curves. Note that the lines are very close, so even small deviations can lead to misclassification. Hence, even an accuracy of 50\% is telling.

Table \ref{tab:appearance_count}  reports only successful executions of NeuroSAT. Similar to the support, in failed executions, the concept is not learned as well. Details omitted.

\begin{table}
  \centering
  \begin{tabular}{ |c |c |c |c |c |}
    \hline
    Dataset   & $i=1$ &$i=2$ & $i=3,4,5$ & $i \ge 6$  \\
    \hline
    SPARSE  &   91$\pm 3$\%  & 77$\pm 8$\% & 85$\pm 6$\% & 85$\pm 10$\% \\
    \hline
    DENSE  &  83$\pm 10$\% & 49$\pm 15$\% & 65$\pm 17$\% & 76$\pm 14$\%\\
    \hline
    PLANTED  & 96$\pm 6$\% & 86$\pm 14$\% & 88$\pm 10$\% & 96$\pm 3$\% \\
    \hline

  \end{tabular}
  \caption{Appearnce count concept. 
Percentage of variables that appear $i$ times and are closest to the corresponding regression line.}
  \label{tab:appearance_count}
\end{table}

\section{Simplification and teachability}\label{sec:teachability}
In this section, we show that the concepts carry over to two simpler architectures of NeruoSAT. In fact, the second, simpler architecture of the two, was untrainable using the cross-entropy loss like NeuorSAT, but was trained with a concept-defined loss. This latter point demonstrates the teachability aspect of the learned concepts.

\subsection{Replacing LSTM with a simpler RNN}\label{sec:RNNvsLSTM}
Our first result shows that the concepts carry over to a simpler, still successful, NN architecture. We replaced the LSTMs $\mathbf{L_u}$ and $\mathbf{C_u}$ in Eq. \eqref{eq:ClauseUpdate} and \eqref{eq:LiteralUpdate} with a single layer RNN $:= \mathbf{th} (FC(input)+FC(history))$, reducing the number of non-linear activations from 5 to 1 per equation. $FC$ is a fully connected layer and $\mathbf{th}=\tanh$ activation function. The new NN is: 
\begin{align}\label{eq:new_RNN}
   & (C^{t+1}, C_h^{t+1}) \leftarrow \mathbf{th} \left[FC\left(M^\top \mathbf{L_{msg}} (L^t)\right)+FC(C_h^t)\right]\\&
    (L^{t+1}, L_h^{t+1}) \leftarrow  \mathbf{th} \left[ FC\left(Flip(L^t), M \mathbf{C_{msg}}(C^t)\right) + FC(L_h^t)\right]. \notag
\end{align}

We re-trained the new architecture using the same train set as NeuroSAT and the same loss function. We found that the accuracy reported in Table \ref{table: performance} is identical, and we replicated the results of Section \ref{sec:concepts}. The details can be found in Tables \ref{tab:consistency sparse}--\ref{tab:appearance_count sparse} in the Appendix.
Figure \ref{FGR: RNN-NeuroSAT} in the Appendix portrays the concept of support using the new RNNs; the results resemble Figure \ref{FGR: support}.  This result shows the concepts' robustness to architectural changes.

\subsection{Further simplification via disentanglement}\label{sec:techaer_student}
We further simplified the RNN in Eq.~\eqref{eq:new_RNN} by removing the history hidden state $C_h^t$ in the clause update message, and the dependency on $L^t$ in the variable update, arriving at 

\begin{align}\label{eq:new_RNN_Detang}
   & C^{t+1} \longleftarrow \mathbf{th}(FC( M^\top L^t))
\\&
   L^{t+1} \longleftarrow \mathbf{th}(FC(M C^t) + FC(L^t - Flip(L^t))). \notag
\end{align}

We refer to this version as DetangledNeuroSAT, as the clause embedding is independent of its prior state, eliminating the need for recurrence. Additionally, instead of concatenating the literal with its negation, we now subtract it.

Interestingly, we could not train DetangledNeuroSAT successfully with the cross-entropy loss function of predicting satisfiability. Thus in some sense, Eq.~\eqref{eq:new_RNN} is the simplest version of NeuorSAT that can be trained with that loss.

However, we successfully trained the architecture described in Equation \eqref{eq:new_RNN_Detang} using a teacher-student approach aimed at reconstructing the embedding of NeuroSAT. 
Ideally, we would have forced the two leading PCs of the teacher and the student to be the same (or even just the sparse PCs, 16 dimensions each and 64 dim in total for the student and the teacher for the top two PCs). However, PCA is not a differentiable operator, and hence we needed a different way to encode the idea of a ``similar reduced dimensionality".

We augmented each of NeuroSAT (the teacher) and DetangledNeuroSAT (the student) with two linear MLPs without activation functions, configured as an encoder-decoder  (128dim $\longrightarrow$ 64dim $\longrightarrow$ 128dim). One MLP for the literals' embedding and one for the clauses'. The teacher was defined as the encoder's output of NeuroSAT's embedding and, similarly, the student. We used the MSE between the encoding outputs of the student and the teacher as a loss function. Each training step was composed of two substeps, one when freezing $L^{(t)}$ and updating the gradient according to the loss over $C^{(t)}$ and vice versa. 
This was repeated for 50 iterations $t=1,\ldots,50$. 

Interestingly, the reduction of dimensionality to 64 was necessary for learning (taking the loss over the 128-dim embedding did not converge).

As with Eq.~\eqref{eq:new_RNN}, the accuracy reported in Table \ref{table: performance} is identical for DetangledNeuroSAT, and we replicated the results of Section \ref{sec:concepts}. The details can be found in Tables \ref{tab:consistency sparse}--\ref{tab:appearance_count sparse} in the Appendix. Figure \ref{FGR: Detangled-NeuroSAT} portrays the concept of support using DetangledNeuroSAT; the results resemble Figure \ref{FGR: support}. Any further subtraction or simplification of Eq.\eqref{eq:new_RNN_Detang} resulted in failed training.

\section{NeuroSAT as a texbook algorithm}\label{sec: textbook}
Using all the concepts we identified for NeuroSAT, we  write it as a textbook algorithm. The description given in Algorithm 1 should not be understood as an exact mirror of the execution of NeuroSAT, but rather as a pseudo-code that captures the guiding principles of the execution.

\begin{figure}[h!]
\centering
        \includegraphics[ width=\columnwidth ]{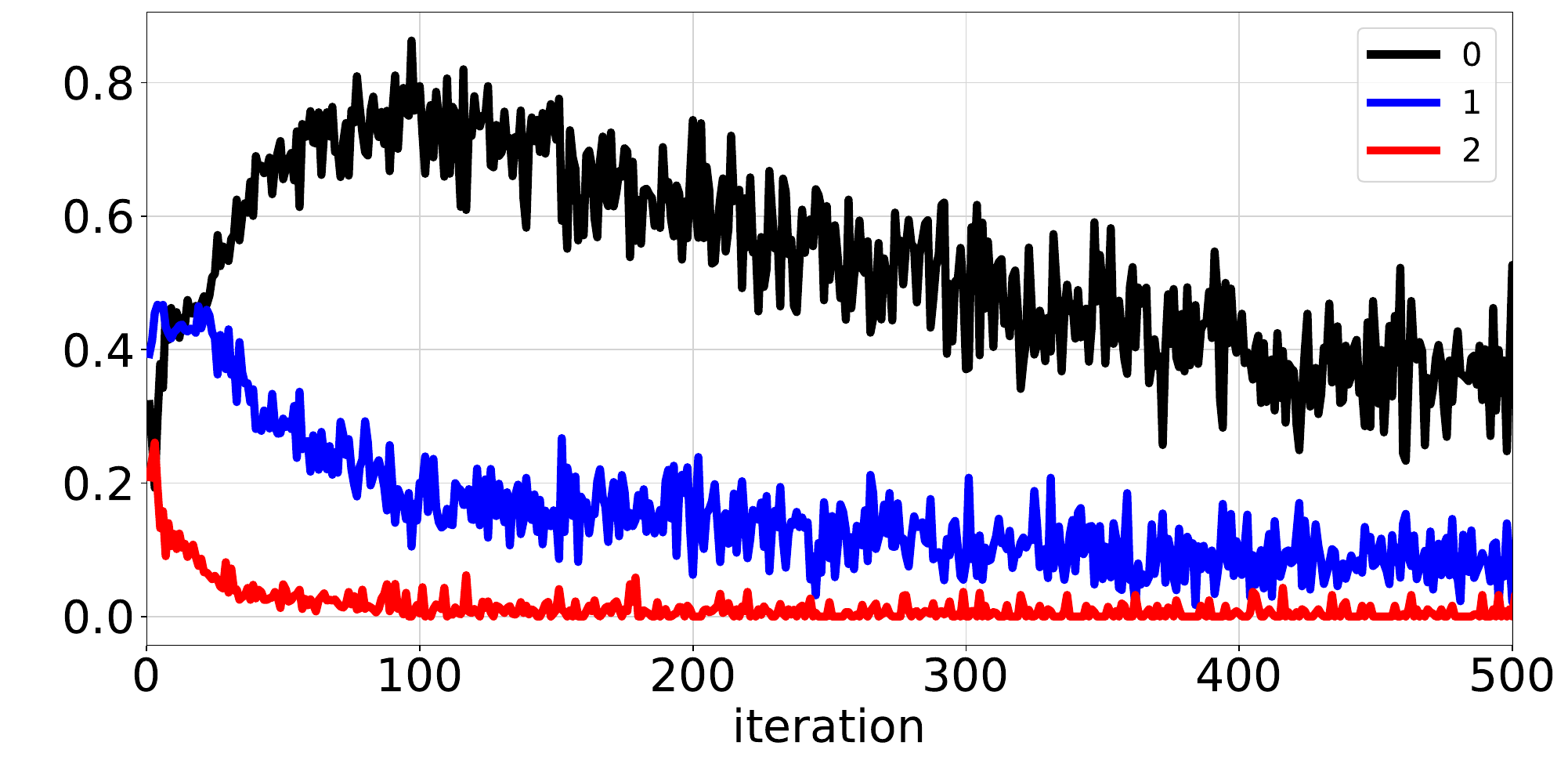}
        \caption{
        The flip probability $p_{ij}$ for a variable with support $i=0,1,2$ in iteration $j$; $p_{ij}$ increases as support decreases. This pattern, averaged over 100 runs with c=4.1,n=1,500, holds for all c’s in DENSE.
        \\
        }
        \label{FGR: support_iter_si}
\end{figure}

Before we present the algorithm, we list a few more facts that we verified: (1) NeuroSAT learns all the concepts by iteration at most 30 for all datasets (2)  NeuroSAT only changes the embedding of literals whose embedding has PC1 value in the range $[-2,2]$; According to the ranges explored in Table \ref{tab:Literal_support} this translates to the distribution in  Figure \ref{FGR: support_iter_si} (3) From iteration 30 onwards, NeuroSAT changes in every iteration the assignment of less than two variables on average. 

    \begin{algorithm}
    \caption{Textbook NeuroSAT}
    \label{alg: neurosat}
    \begin{algorithmic}[1]
    \State $\phi_1 \leftarrow MAJ$ (iteration $t=1$)
    \State Compute the support for all variables (iterations $t=1\ldots 30$)
    \For{$t=31 \ldots 500$}
        \State Sample $x$ according to the distribution in Figure \ref{FGR: support_iter_si}
        \State Set $\phi_{t+1}(x) \gets \neg\phi_t(x)$
    \EndFor
    
     \State \Return $\phi_{500}$ derived by the assignment concept (Section \ref{SEC: con_ass})
    \end{algorithmic}
    \end{algorithm}

\begin{figure}[h!]
    \centering
        \includegraphics[ width=\columnwidth ]{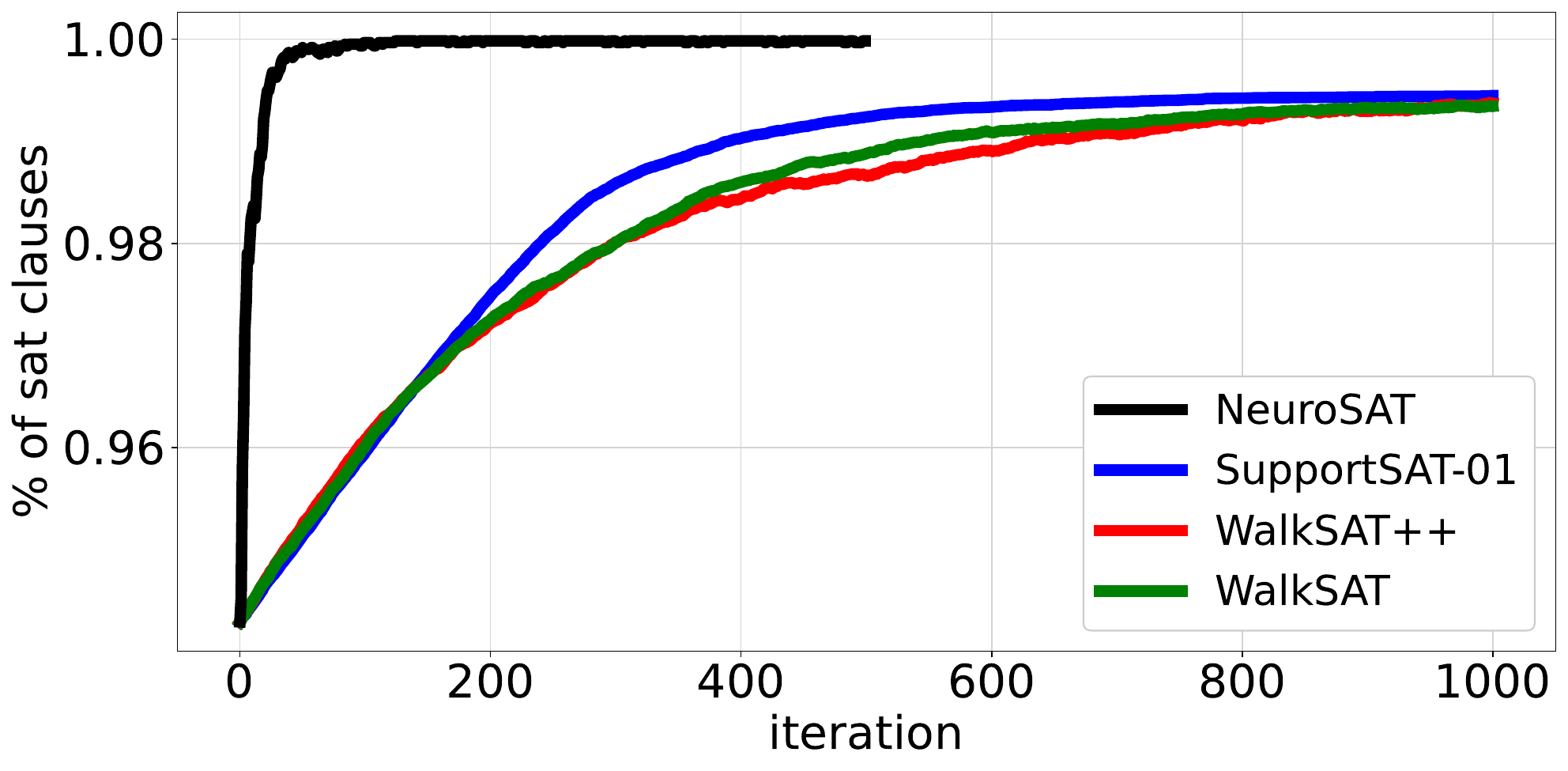}
        \caption{SupportSAT-01 improves WalkSAT. NeuroSAT converges the fastest, then SupportSAT-01, WalkSAT++ and finally WalkSAT. All algorithms start from MAJ. Averaged over 40 instances with $c=4.1, n=1500$.}
    \label{FGR: walks}
\end{figure}

A few observations about this write-up of NeuroSAT:
\begin{itemize}
    \item In line 2, while the concepts are being figured out in the first 30 iterations, NeuroSAT also flips the assignment of tens of variables per iteration. While doing so, NeuroSAT gets closer to the satisfying assignment, as can be seen in Figure \ref{FGR: walks}. We don't have any claim about what principles guide NeuorSAT in that stage.
    \item Recall the WalkSAT concept, Definition \ref{defn:Walksat} and compare to the for-loop in lines 3-6. Indeed NeuroSAT performs a variant of WalkSAT with flip probabilities according to the empirical distribution presented in Figure \ref{FGR: support_iter_si}.
\end{itemize}

Inspired by NeuroSAT’s utilization of the support measure (Figure \ref{FGR: support_iter_si}), we tweak WalkSAT \`a la NeuroSAT. We call the new version SupportSAT-01, and its rule is as follows. With probability 2/3,  flip a random variable with support 0 that appears in some UNSAT clause; with probability 1/3, flip a variable with support 1. 
For comparison, we run WalkSAT with parameter $p=0.1$ (the probability of making a random move); to fix $p$ we ran a grid search over $(0,1)$, with steps of 0.1. We also compare to WalkSAT++, a variant taken from \cite{walksat_proof}, which always flips variables with support 0 if possible, and otherwise it runs a WalkSAT move. \textcolor{black}{The asymptotical running time of all three versions is identical, $O(T\cdot(m+n))$, where $m$ is the number of clause, $n$ the number of variables and $T$ the maximum number of iterations.}

Figure \ref{FGR: walks} shows that SupportSAT-01 converges faster to a satisfying assignment, next WalkSAT++ and last WalkSAT.

\section{Conclusion } \label{SEC: conclusion}
In this work, we have extended the concept-learning paradigm to combinatorial optimization problems, focusing on the satisfiability (SAT) problem. Our study of NeuroSAT, a graph neural network trained to solve SAT, reveals that it inherently learns several key algorithmic concepts. These concepts, which include assignment consistency, support, backbone variables, majority vote, and appearance count, are critical for understanding and solving SAT instances. 

We demonstrated that these concepts are encoded in the leading principal components of the network's latent space activations. By analyzing the covariance matrix of the embeddings, we discovered that concepts like support and backbone variables are crucial for NeuroSAT's performance. Our experiments showed that NeuroSAT's concept-learning capabilities enable it to find satisfying assignments efficiently and adapt to various out-of-distribution SAT problems, including the planted and backbone datasets.

Moreover, we illustrated the three anchors of concept definition: usefulness in solving the task; Teachability of these concepts by training simpler neural network architectures using a concept-defined loss function; Minimality by using sparse PCA while retaining the same performance statistics. 

We further showcased the practical applications of concept learning in explainable AI by rewriting NeuroSAT as a textbook algorithm and enhancing classical SAT-solving heuristics, like WalkSAT, with NeuroSAT-inspired modifications.

In this work, the concepts were excavated manually by computing the principal components, reducing dimensionality, and using domain expertise to check various salient SAT-solving concepts. Our work should be seen as a proof of concept (no pun intended) that concept learning is possible for complex algorithmic tasks.  {\color{black} Some elements, such as PCA, regression analysis, Spearman rank correlation, and distillation, can indeed be automated. However, while automating the extraction of potential concepts is feasible, the ultimate purpose of concept learning is to understand how the neural network functions, thereby increasing confidence in its decisions. Thus, human interpretation remains essential, as a human will need to assign meaningful labels to concepts—much like assigning labels to regions in images.}

The next challenge is developing a structured framework and language to automate concept excavation effectively. Questions around automating dataset selection, choosing comparison algorithms, and identifying relevant concepts for correlation remain open for future exploration. Encouraging progress, as seen in \cite{kim2022beyond,schut2023bridging}, provides optimism for advancing these tasks further.

\section{Declaration of Generative AI and AI-assisted technologies in the writing process } \label{SEC: declaration}
Statement: During the preparation of this work, the author(s) used ChatGPT in order to minor edit sentences. After using this tool/service, the author(s) reviewed and edited the content as needed and take(s) full responsibility for the content of the publication.

\bibliographystyle{unsrt}
\bibliography{main}

\newpage
\appendix
\section{}\label{appendix}

  \begin{figure}[h]
        \centering
        \includegraphics[angle=0,width=0.4\textwidth]{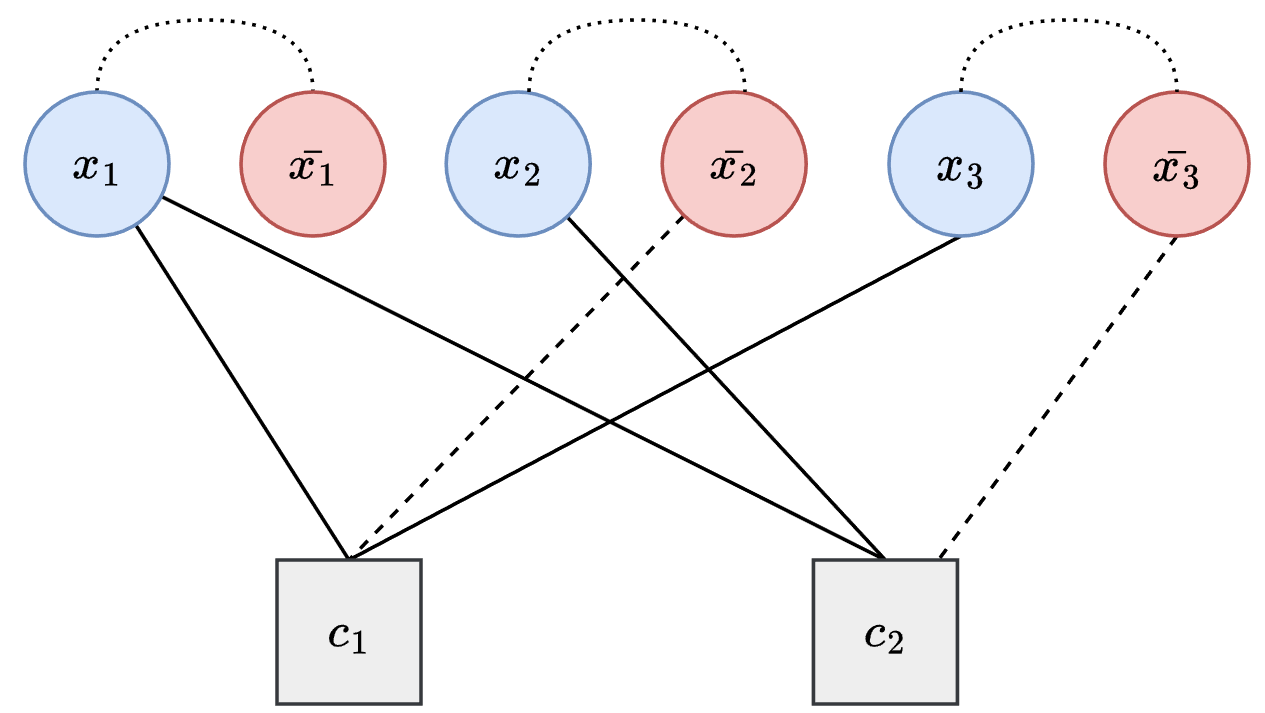}
        \caption{The instance $F=(x_1 \vee \bar{x_2} \vee x_3) \land (x_1 \vee x_2 \vee \bar{x_3})$ is transformed into an input graph for NeuroSAT. A node is created for each clause and literal. Then,  clause nodes are connected to the variables appearing in them, and  each literal to its negation.}
        \label{FGR: sat_to_graph}
    \end{figure}

    \begin{figure}[h]
        \centering
            \includegraphics[angle=0,width=0.4\textwidth]{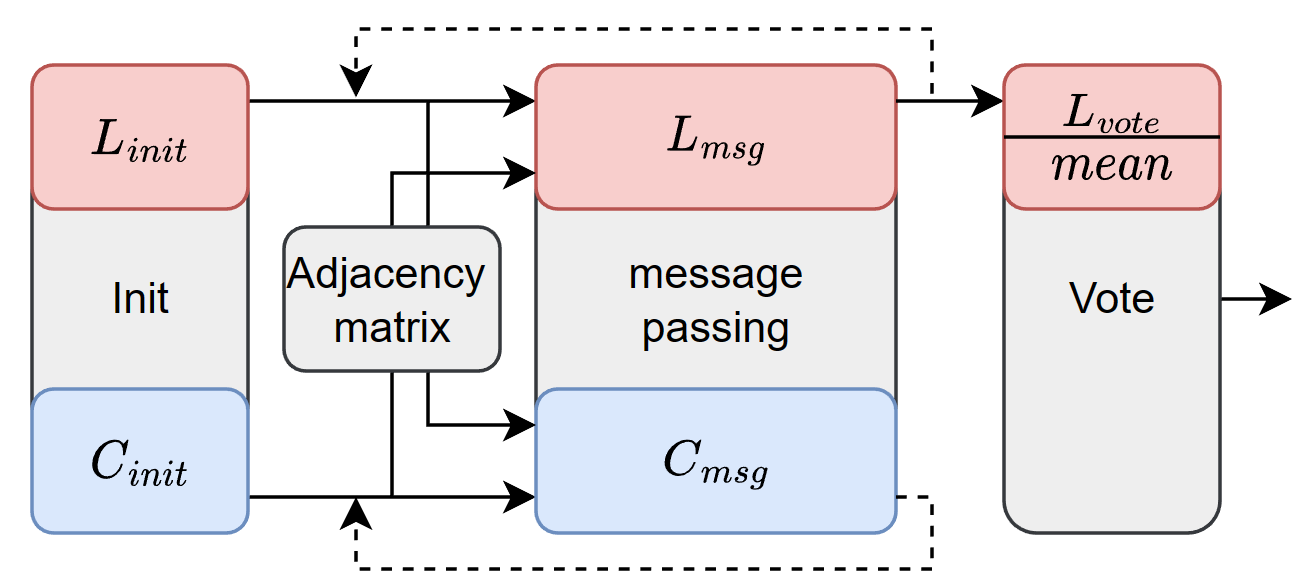}
        \caption{
        A high-level view of the NeuroSAT architecture. NeuroSAT can be viewed as a message-passing algorithm, alternating between clause messages, $C$, and literal messages, $L$. The update rules is $(C^{(t+1)}, C_h^{(t+1)}) \leftarrow C_u([C_h^{(t)}, M L_{msg}(L^{(t)})])$ for the clauses, and 
            $(L^{(t+1)}, L_h^{(t+1)}) \leftarrow L_u([L_h^{(t)}, FLIP(L^{(t)}), M C_{msg}(C^{(t+1)})])$ for the literals. 
            After $T$ iterations, compute $L_*^{(T)} \leftarrow L_{vote}(L^{(T)})$, and lastly to achieve the SAT-UNSAT prediction $y^{(T)} \leftarrow mean(L_*^{(T)})$.
        }
        \label{FGR: full_model}
    \end{figure}

\begin{figure}
    \centering
        \begin{subfigure}[t]{1\columnwidth }
            \centering
            \includegraphics[ width=\textwidth]{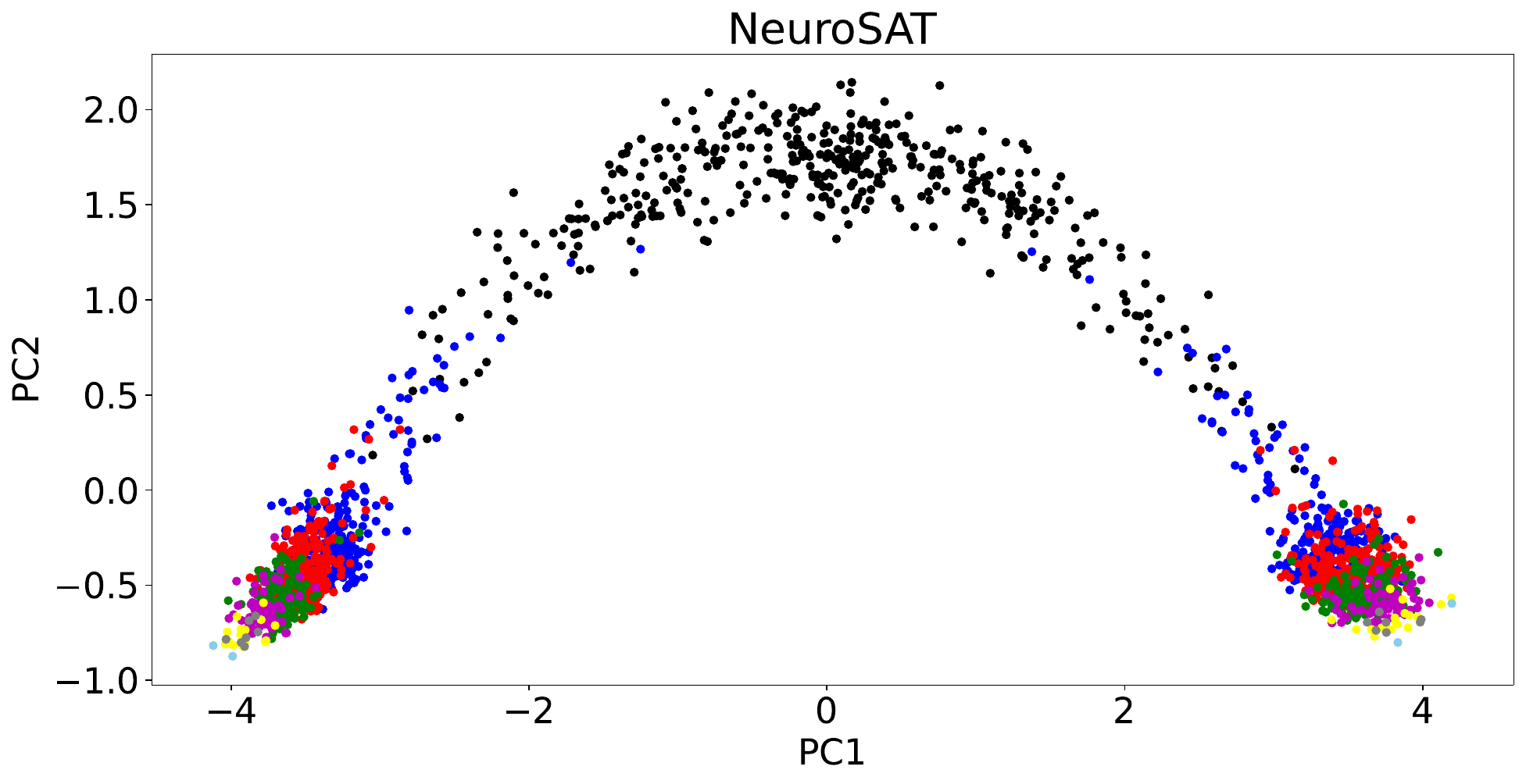}
             \subcaption{NeuroSAT}
             \label{FGR: NeuroSAT}
        \end{subfigure}
        \hfill
                \begin{subfigure}[t]{1\columnwidth }
            \centering
            \includegraphics[ width=\textwidth]{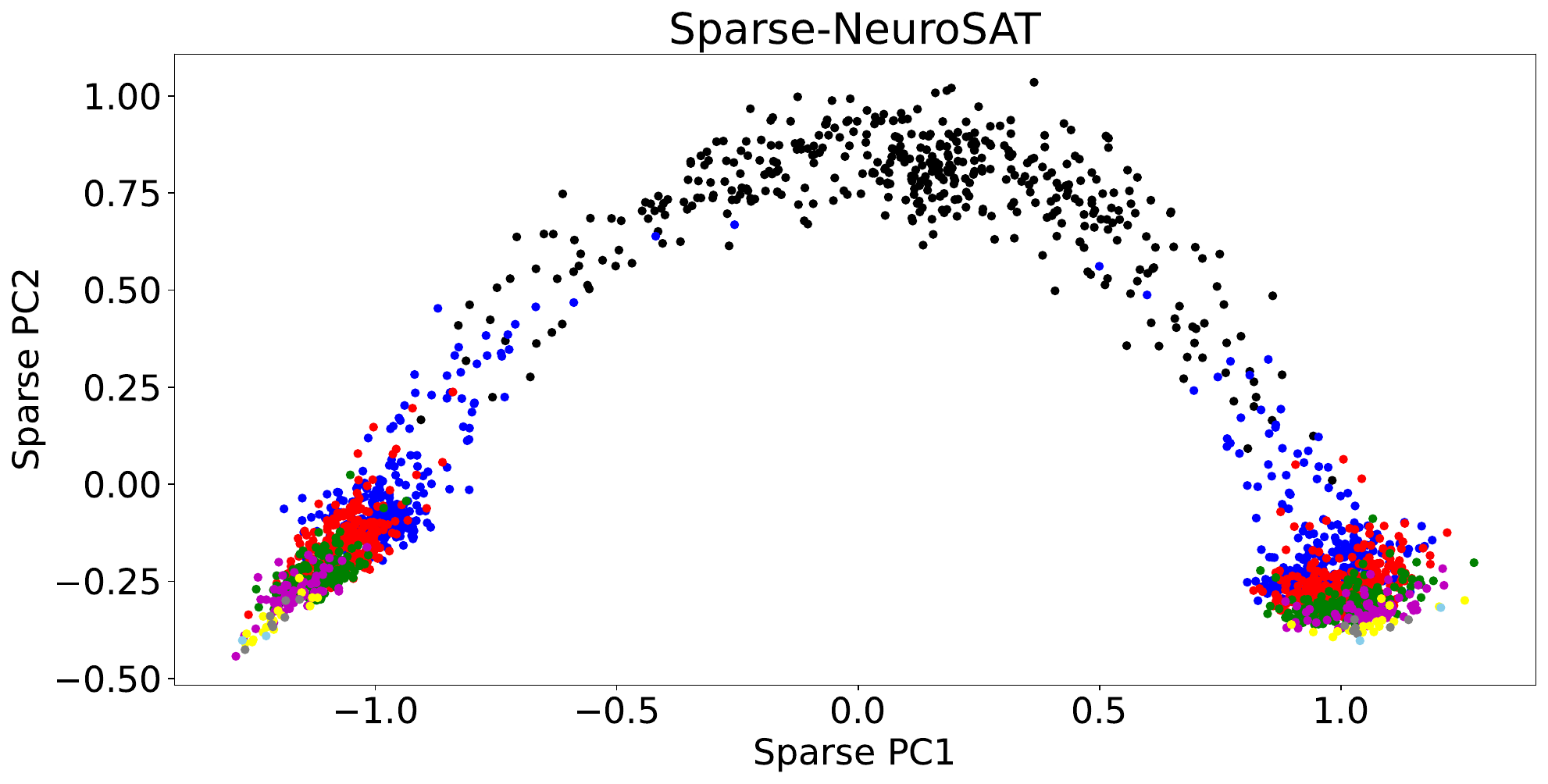}
             \subcaption{Sparse-NeuroSAT}
             \label{FGR: Sparse-NeuroSAT}
        \end{subfigure}
        \hfill
                \begin{subfigure}[t]{1\columnwidth }
            \centering
            \includegraphics[ width=\textwidth]{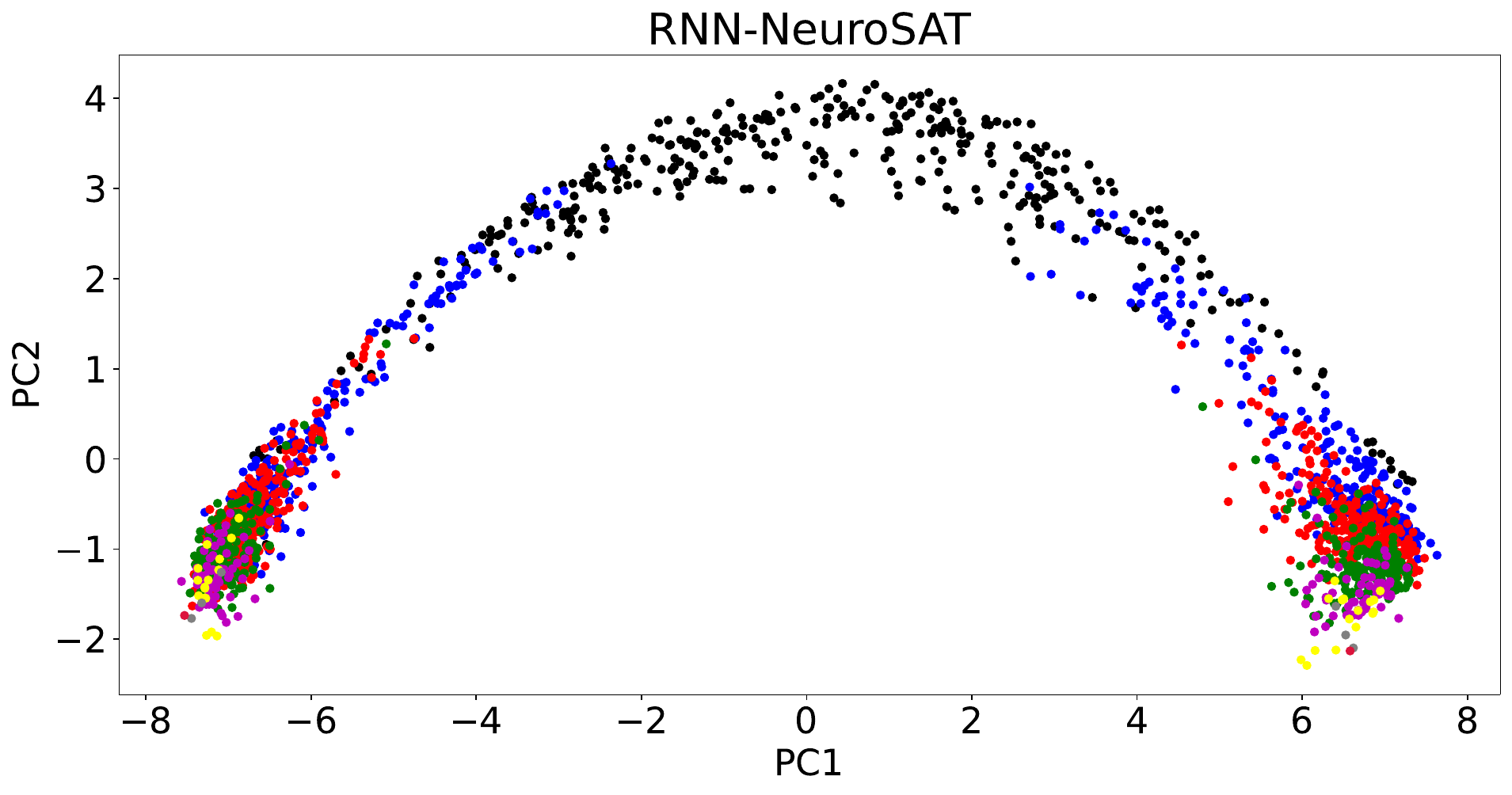}
             \subcaption{RNN-NeuroSAT}
             \label{FGR: RNN-NeuroSAT}
        \end{subfigure}
        \hfill
            \begin{subfigure}[t]{1\columnwidth }
            \centering
            \includegraphics[ width=\textwidth]{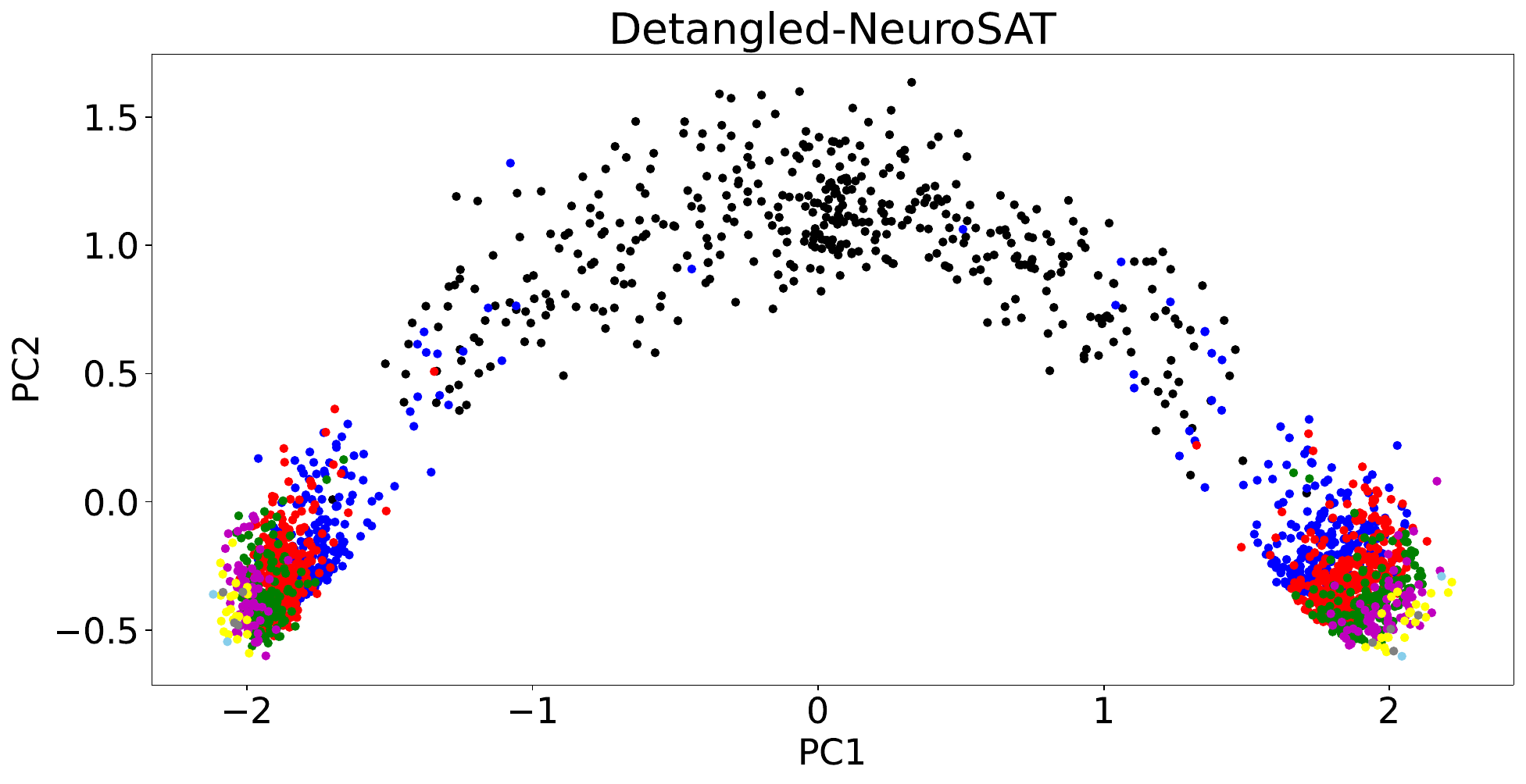}
             \subcaption{Detangled-NeuroSAT}
             \label{FGR: Detangled-NeuroSAT}
        \end{subfigure}
    \caption{Literlas' embedding on different versions of NeuroSAT (NeuroSAT, Sparse-NeuroSAT, RNN-NeuroSAT, Detangled-NeuroSAT). All show an general arch shape with the concept of support learned.} 
    \label{FGR: support_versions}
    \end{figure}

\begin{table*}
  \centering
  \begin{tabular}{ |c| c| c| c| }
  \hline
     Dataset & Model & \multicolumn{2}{|c|}{\% of contradictions} \\
    \cline{3-4} 
    &  &  Full cov &  Sparse cov   \\
    \hline
     & Original & 7.3 $\pm 2.7$ \% &6.4 $\pm 1.2$ \% \\ 
    \cline{2-4} 
    SPARSE&  RNN & 3.4 $\pm 2.1$ \% & 6.5 $\pm 3.3$ \%\\ 
    \cline{2-4} 
    &  Detangled& 4.3 $\pm 2.7$ \% & 4.8 $\pm 1.4$ \%\\ 
    \hline
     &Original & 3.2 $\pm 1.5$ \% &  4 $\pm 1$ \%\\ 
    \cline{2-4} 
    DENSE& RNN & 4.2 $\pm 2$ \% & 1 $\pm 0.4$ \% \\ 
    \cline{2-4} 
     & Detangled & 3.2 $\pm 1.5$ \% & 1.9 $\pm 0.7$ \% \\ 
    \hline
     & Original & 2 $\pm 1$\% & 2 $\pm 1$ \%\\ 
    \cline{2-4} 
    PLANTED & RNN & 2 $\pm 1$\% & 0.2 $\pm 0.1$ \% \\ 
    \cline{2-4} 
     & Detangled & 2 $\pm 1$\% & 1 $\pm 0.4$ \% \\ 
    \hline
  \end{tabular}
  \caption{Assignment Consistency Rates. The average \% of contradictions per iteration over all instances in the dataset.}
  \label{tab:consistency sparse}
\end{table*}

\begin{table*}
\centering
\begin{tabular}{ |c| c|  c|  c | c | c | c  |c  |c | c|  } 
 \hline
Covariance Matrix & Distribution & Model &  $c<1.6$ &  $c<3.75$ &  $c=3.8$  &  $c=4.0$ &  $c=4.1$ &  $c=4.2$ &  $c=4.25$   \\
 \hline
&& Original  & $100\%$ & $100\%$ & $100\%$  & $67\%$ & $48\%$ & $58\%$ &  $43\%$ \\
&RANDOM &RNN  & $100\%$ & $100\%$ & $96\%$ & $70\%$ & $32\%$ & $63\%$ &  $26\%$ \\
Full&&Detangled  & $100\%$  & $84\%$ & $75\%$ & $45\%$ &  $32\%$ &  $23\%$ &  $16\%$ \\
\cline{2-10}
&&Original & $100\%$ &  $100\%$ & $100\%$ & $100\%$ & $100\%$  & $100\%$  & $100\%$  \\
&PLANTED& RNN & $97\%$ &  $99\%$ & $100\%$ & $100\%$ & $100\%$ & $100\%$ & $100\%$ \\
&&Detangled & $100\%$ &  $100\%$ & $100\%$ & $100\%$ & $100\%$ & $100\%$ & $96\%$  \\

 \hline
 \hline
 && Original  & $100\%$  & $100\%$ & $96\%$ & $67\%$ &  $48\%$ &  $58\%$ &  $43\%$ \\
&RANDOM &RNN  & $100\%$ & $96\%$ & $90\%$ & $70\%$ & $32\%$ & $63\%$ &  $23\%$ \\
Sparse&&Detangled  & $100\%$  & $91\%$ & $70\%$ & $40\%$ &  $32\%$ &  $23\%$ &  $16\%$ \\
\cline{2-10}
&&Original & $100\%$ &  $100\%$ & $100\%$ & $100\%$ & $100\%$ & $100\%$ & $100\%$  \\
&PLANTED& RNN & $98\%$ &  $100\%$ & $100\%$ & $100\%$ & $100\%$ & $100\%$ & $100\%$ \\
&&Detangled & $100\%$ &  $100\%$ & $100\%$ & $100\%$ & $100\%$ & $100\%$ & $100\%$  \\

\hline
\end{tabular}
\caption{The average rate of Sparse NeuroSAT finding a satisfying assignment.}
\label{table: performance sparse}
\end{table*}

\begin{table*}
  \centering
  \begin{tabular}{ |c| c| c| c| }
    \hline
     Dataset & Model & \multicolumn{2}{|c|}{\% of concept-abiding clauses} \\
    \cline{3-4} 
    &  &  Full cov &  Sparse cov   \\
    \hline
    & Original & $61 (\pm 8)$ \% & $61 (\pm 6)$ \% \\ 
    \cline{2-4} 
    SPARSE & RNN & $68 (\pm 3)$ \%  & $68 (\pm 3)$ \% \\ 
    \cline{2-4} 
     & Detangled & $62 (\pm 6.5)$ \% & $61 (\pm 6)$ \% \\ 
    \hline
    
     & Original & $89 (\pm 6)$ \% & $87 (\pm 7)$ \% \\ 
    \cline{2-4} 
    DENSE & RNN & $89 (\pm 7)$ \% & $89 (\pm 7)$ \% \\ 
    \cline{2-4} 
     & Detangled & $89 (\pm 6)$ \% & $88 (\pm 6)$ \% \\ 
    \hline
     & Original & $97 (\pm 5)$ \% & $97 (\pm 5)$ \% \\ 
    \cline{2-4} 
    PLANTED & RNN & $98 (\pm 5)$ \% & $98 (\pm 5)$ \% \\ 
    \cline{2-4} 
     & Detangled & $97 (\pm 5)$ \% & $97 (\pm 5.2)$ \% \\ 
    \hline
  \end{tabular}
  \caption{Clause support concept. The average \% of concept-abiding support clauses. }
  \label{tab:Clause_support sparse}
\end{table*}

\begin{table*}
  \centering
  \begin{tabular}{ |c |c |c |c |c |c |c |}
    \hline
    Covariance Matrix &Model& Dataset  & Support 0 & Support 1 & Support 2 & Support 3  \\
    \hline
    &&SPARSE  & $\pm[0, 2]$  & $\pm[2.1, 3.3]$ & $\pm[2.7, 3.5]$  &  $\pm[3, 3.7]$  \\
    &&&94\%&80\%&86\%&80\%\\
    \cline{3-7}
    &&DENSE &  $\pm[0, 1.5]$  & $\pm[1.6, 2.8]$ & $\pm[2.4, 3.0]$  &  $\pm[2.6, 3.2]$  \\
    &&&81\%&76\%&84\%&92\%\\
    \cline{3-7}
   &Original &DENSE &  $\pm[0, 1.5]$  & $\pm[1.6, 2.8]$ & $\pm[2.4, 3.0]$  &  $\pm[2.6, 3.2]$  \\
    &&FAILED & 46\% & 60\% & 70\% & 76\% \\
    \cline{3-7}
   &&DENSE  &  $\pm[0, 1.5]$  & $\pm[1.6, 2.8]$ & $\pm[2.4, 3.0]$  &  $\pm[2.6, 3.2]$  \\
    &&FAR-FAILED & 60\% & 43\% & 54\% & 61\% \\
    
    \cline{2-7}
    &&SPARSE & $\pm[0, 0.325]$  & $\pm[0.275, 0.335]$ & $\pm[0.28, 0.35]$  &  $\pm[0.29, 0.4]$  \\
    &&&88\%&76\%&81\%&82\%\\
    \cline{3-7}
    &&DENSE &  $\pm[0, 4.5]$  & $\pm[4.3, 6.3]$ & $\pm[5.5, 6.5]$  &  $\pm[5.9, 6.8]$  \\
    &&&80\%&64\%&77\%&79\%\\
    \cline{3-7}
   Full & RNN&DENSE  &  $\pm[0, 4.5]$  & $\pm[4.3, 6.3]$ & $\pm[5.5, 6.5]$  &  $\pm[5.9, 6.8]$  \\
    &&FAILED & 49\% & 65\% & 61\% & 48\% \\
    \cline{3-7}
   &&DENSE &  $\pm[0, 4.5]$  & $\pm[4.3, 6.3]$ & $\pm[5.5, 6.5]$  &  $\pm[5.9, 6.8]$  \\
    &&FAR-FAILED &95\% & 6\% & 1\% & 0.4\% \\
    \cline{2-7}

    &&SPARSE & $\pm[0, 1.0]$  & $\pm[0.9, 1.6]$ & $\pm[1.55, 1.8]$  &  $\pm[1.73, 2.0]$  \\
    &&&90\%&71\%&71\%&72\%\\
    \cline{3-7}
    &&DENSE &  $\pm[0, 1.2]$  & $\pm[1.0, 1.85]$ & $\pm[1.65, 1.95]$  &  $\pm[1.8, 2.2]$  \\
    &&&88\%&84\%&82\%&80\%\\
    \cline{3-7}
   & Detangled&DENSE &  $\pm[0, 1.2]$  & $\pm[1.0, 1.85]$ & $\pm[1.65, 1.95]$  &  $\pm[1.8, 2.2]$  \\
    &&FAILED & 46\% & 0\% & 65\% & 73\% \\
    \cline{3-7}
   &&DENSE &  $\pm[0, 1.2]$  & $\pm[1.0, 1.85]$ & $\pm[1.65, 1.95]$  &  $\pm[1.8, 2.2]$  \\
    &&FAR-FAILED & 57\% & 0\% & 36\% & 50\% \\
    \hline
    \hline

    &&SPARSE  & $\pm[0, 0.45]$  & $\pm[0.39, 0.85]$ & $\pm[0.87, 1.2]$  &  $\pm[3, 3.7]$  \\
    &&&85\%&77\%&74\%&95\%\\
    \cline{3-7}
    &&DENSE &  $\pm[0, 0.5]$  & $\pm[0.3, 0.87]$ & $\pm[0.75, 0.92]$  &  $\pm[0.8, 1.0]$  \\
    &&&98\%&81\%&75\%&80\%\\
    \cline{3-7}
   &Original &DENSE &  $\pm[0, 0.5]$  & $\pm[0.3, 0.87]$ & $\pm[0.75, 0.92]$  &  $\pm[0.8, 1.0]$  \\
    &&FAILED &22\% & 53\% & 51\% & 58\% \\
    \cline{3-7}
   &&DENSE  &   $\pm[0, 0.5]$  & $\pm[0.3, 0.87]$ & $\pm[0.75, 0.92]$  &  $\pm[0.8, 1.0]$  \\
    &&FAR-FAILED& 17\% & 50\% & 54\% & 63\% \\
    
    \cline{2-7}
    &&SPARSE & $\pm[0, 0.09]$  & $\pm[0.072, 0.093]$ & $\pm[0.08,0.097]$  &  $\pm[0.082,0.1]$  \\
    &&&80\%&78\%&87\%&83\%\\
    \cline{3-7}
    &&DENSE& $\pm[0,1.7]$  & $\pm[1.6, 2.05]$ & $\pm[1.92,2.1]$  &  $\pm[1.97, 2.15]$  \\
    &&&80\%&81\%&88\%&87\%\\
    \cline{3-7}
   Sparse & RNN&DENSE  & $\pm[0,1.7]$  & $\pm[1.6, 2.05]$ & $\pm[1.92,2.1]$  &  $\pm[1.97, 2.15]$  \\
    &&FAILED & 79\%&80\%&74\%&70\%\\
    \cline{3-7}
   &&DENSE& $\pm[0,1.7]$  & $\pm[1.6, 2.05]$ & $\pm[1.92,2.1]$  &  $\pm[1.97, 2.15]$  \\
    &&FAR-FAILED &95\%&21\%&14\%&14\%\\
    \cline{2-7}

    &&SPARSE & $\pm[0,0.8]$  & $\pm[0.9, 1.4]$ & $\pm[1.1,1.5]$  &  $\pm[1.3,1.8]$  \\
    &&&90\%&81\%&92\%&78\%\\
    \cline{3-7}
    &&DENSE & $\pm[0,1.1]$  & $\pm[1.2, 1.6]$ & $\pm[1.3,1.7]$  &  $\pm[1.4,1.75]$  \\
    &&&88\%&80\%&89\%&88\%\\
    \cline{3-7}
   & Detangled&DENSE & $\pm[0,1.1]$  & $\pm[1.2, 1.6]$ & $\pm[1.3,1.7]$  &  $\pm[1.4,1.75]$  \\
    &&FAILED &18\%&45\%&63\%&67\%\\
    \cline{3-7}
   &&DENSE & $\pm[0,1.1]$  & $\pm[1.2, 1.6]$ & $\pm[1.3,1.7]$  &  $\pm[1.4,1.75]$  \\
    &&FAR-FAILED & 17\%&42\%&59\%&63\%\\
    \hline
    \hline

  \end{tabular}
  \caption{Literal support concept. $PC1$ range and the percentage of clauses with support $i$ that fall in that range. We shorthand $\pm[a,b]$ for $[-a,-b] \cup [a,b].$}
  \label{tab:Literal_support sparse}
\end{table*}

\begin{table*}
  \centering
  \begin{tabular}{ |c| c |c |c |}
    \hline
     Dataset & Model & \multicolumn{2}{|c|}{Hamming(MAJ,$\phi_1$)} \\
    \cline{3-4} 
    &  &  Full cov &  Sparse cov   \\
    \hline 
     & Original &  $0.0 \pm 0.0$ & $0.14 \pm 0.03$ \\
    \cline{2-4} 
    SPARSE & RNN &  $0.0 \pm 0.0$ &  $14.5 \pm 2.9$ \\
    \cline{2-4}
     & Detangled &  $0.005 \pm 0.005$ &  $14.9 \pm 3.1$ \\
     \cline{1-4}
     \cline{1-4}
    \hline
     & Original    & $0.0 \pm 0.0$ &   $0.20 \pm 0.013$ \\
    \cline{2-4} 
    DENSE  & RNN  & $0.0 \pm 0.0$ &  $20 \pm 1.6$ \\
    \cline{2-4}
     & Detangled   & $0.0 \pm 0.0$ &  $19.4 \pm 12.5$ \\
     \cline{1-4}
     \cline{1-4}
    \hline
     &  Original  & $0.00 \pm 0.00$  &  $0.17 \pm 0.014$  \\
    \cline{2-4} 
    PLANTED & RNN  & $0 \pm 0$  &  $18 \pm 1.6$ \\
    \cline{2-4} 
     & Detangled  & $0.0 \pm 0.0$ &  $18 \pm 12.6$ \\
    \hline
  \end{tabular}
  \caption{The Majority Vote concept. The average Hamming distance between MAJ and $\phi_1$, NeuroSAT's output after the 1st iteration.}
  \label{tab:MAJ sparse}
\end{table*}

\begin{table*}
  \centering
  \begin{tabular}{ |c| c| c| c| c |c| c| }
    \hline
    Covariance Matrix & Dataset &Model & $i=1$ &$i=2$ & $i=3,4,5$ & $i \ge 6$ \\
    \hline
     & & Original  &   91 $\pm 3$\%  & 77 $\pm 8$\% & 85 $\pm 6$\% & 85 $\pm 10$\%\\ 
    \cline{3-7}
    & SPARSE & RNN  &   60 $\pm 9$\%  & 50 $\pm 11$\% & 69 $\pm 9$\% & 78 $\pm 14$\% \\
    \cline{3-7}
     & & Detangled  &   82 $\pm 7$\%  & 58 $\pm 11$\% & 79 $\pm 8$\% & 80 $\pm 13$\% \\
    \cline{2-7}
    & & Original  &  83 $\pm 10$\% & 49 $\pm 15$\% & 65 $\pm 17$\% & 76 $\pm 14$\% \\
    \cline{3-7}
    Full & DENSE &  RNN &  57 $\pm 13$\% & 35 $\pm 8$\% & 66 $\pm 12$\% & 87 $\pm 3$\%\\
    \cline{3-7}
    & &Detangled &  65 $\pm 8$\% & 38 $\pm 8$\% & 69 $\pm 6$\% & 83 $\pm 4$\%\\
    \cline{2-7}
    & & Original  & 96 $\pm 6$\% & 86 $\pm 14$\% & 88 $\pm 10$\% & 96 $\pm 3$\% \\
    \cline{3-7}
    & PLANTED& RNN  & 87 $\pm 14$\% & 73 $\pm 25$\% & 85 $\pm 12$\% & 96 $\pm 3$\% \\
    \cline{3-7}
    & &Detangled  & 91 $\pm 11$\% & 81 $\pm 19$\% & 92 $\pm 7$\% & 97 $\pm 2$\% \\
    \hline \hline
    
    & & Original  & 83 $\pm 5$\%  & 60 $\pm 8$\% & 69 $\pm 7$\% & 71 $\pm 16$\% \\
    \cline{3-7}
    & SPARSE & RNN  & 62 $\pm 9$\%  & 51 $\pm 10$\% & 77 $\pm 5$\% & 83 $\pm 4$\% \\
    \cline{3-7}
     & & Detangled  &  81 $\pm 7.3$\%  & 53 $\pm 12.1$\% & 81 $\pm 6.9$\% & 87 $\pm 4.1$\% \\
    \cline{2-7}
    & & Original  &  72 $\pm 15$\% & 39 $\pm 15$\% & 58 $\pm 19$\% & 73 $\pm 14$\%\\
    \cline{3-7}
    Sparse & DENSE &  RNN &   28 $\pm 16$\%  & 12 $\pm 10$\% & 22 $\pm 8$\% & 80 $\pm 5$\% \\
    \cline{3-7}
    & &Detangled &   70 $\pm 10$\%  & 40 $\pm 9$\% & 67 $\pm 6$\% & 86 $\pm 3.8$\% \\
    \cline{2-7}
    & & Original & 82 $\pm 18$\% & 84 $\pm 13$\% & 97 $\pm 3$\% & 92 $\pm 11$\%\\
    \cline{3-7}
    & PLANTED& RNN  &   NA $\pm NA$\%  & NA $\pm NA$\% & 100 $\pm 0$\% & 99.9 $\pm 0.1$\% \\
    \cline{3-7}
    & &Detangled  & NA $\pm NA$\%  & NA $\pm NA$\% & 100 $\pm 0$\% & 99.9 $\pm 0.1$\% \\
    \hline \hline

  \end{tabular}
  \caption{Appearnce count concept. 
Percentage of variables that appear $i$ times and are closest to the corresponding regression line.}
  \label{tab:appearance_count sparse}
\end{table*}

\end{document}